\documentclass[10pt,twocolumn,letterpaper]{article}
\usepackage[accsupp]{axessibility} 
\usepackage{cvpr}              

\usepackage[dvipsnames]{xcolor}
\newcommand{\red}[1]{{\color{red}#1}}

\definecolor{cvprblue}{rgb}{0.21,0.49,0.74}
\usepackage[pagebackref,breaklinks,colorlinks, citecolor=cvprblue]{hyperref}
\usepackage{makecell} 
\usepackage{graphicx}
\usepackage{booktabs}
\usepackage{caption}
\usepackage{xcolor} 
\usepackage{multirow}
\def\paperID{6616} 
\def\confName{CVPR}
\def\confYear{2024}

\title{HashPoint: Accelerated Point Searching and Sampling for Neural Rendering}

\author{Jiahao Ma\textsuperscript{1,2}, Miaomiao Liu\textsuperscript{1}, David Ahmedt-Aristizabal\textsuperscript{2}, Chuong Nguyen\textsuperscript{2} \\
Australian National University\textsuperscript{1}, CSIRO Data61\textsuperscript{2} \\
 {\tt\small \{jiahao.ma, miaomiao.liu\}@anu.edu.au}\\ {\tt\small\{jiahao.ma, david.ahmedtaristizabal, chuong.nguyen\}@data61.csiro.au}
}

\begin{document}

\twocolumn[{%
\renewcommand\twocolumn[1][]{#1}%
\maketitle
\begin{center}
    \centering
    \captionsetup{type=figure}
    
\vspace*{-2.0em}
\centering
    \includegraphics[width=0.96\textwidth]{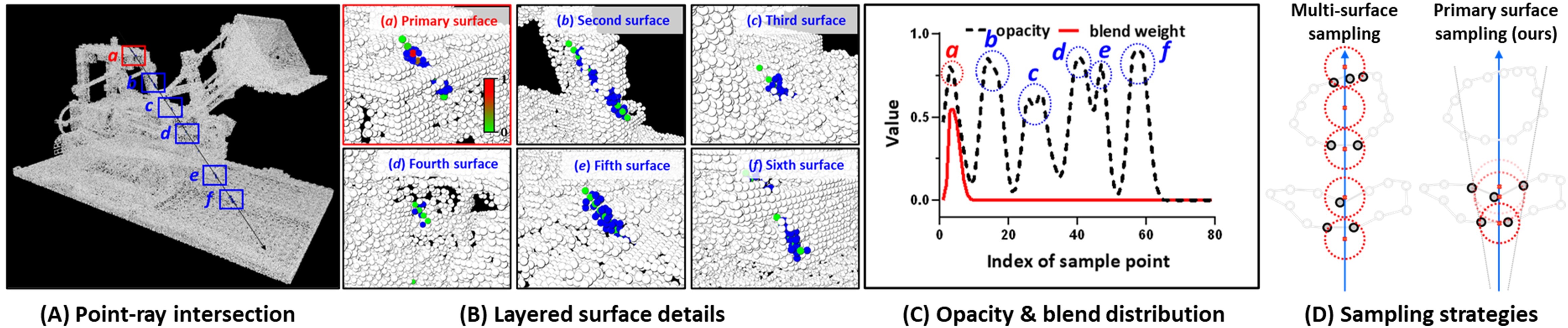}
    \caption{ HashPoint combines rasterization and ray-tracking approaches to optimize point searching and adaptive sampling on primary surfaces. 
    \textbf{(A)} A ray intersects at six surfaces \textit{a}$\sim$\textit{f}. 
    \textbf{(B)} Zoomed-in surfaces to show sample points in \textcolor{green}{green} and \textcolor{red}{red}, with the nearby retrieved point cloud in \textcolor{blue}{blue}. A shift from green to red represents increasing blend weight of the sample points.
   \textbf{(C)} The graph illustrates opacity and rendering weight changes along a camera ray. The first peak, labeled `\textit{a}', indicates the \textit{primary surface} where the ray first hit. Higher rendering weights mean more significant sample points. Notably, the primary surface often dominates the rendering process. 
    \textbf{(D)} Sampling across multiple surfaces versus our HashPoint which accelerates the process via sampling solely on the primary surface.
    } 
    \label{fig:ray_tracing_process}

\end{center}
}]

\begin{abstract}
\vspace{-10pt}
In this paper, we address the problem of efficient point searching and sampling for volume neural rendering. 
Within this realm, two typical approaches are employed: rasterization and ray tracing. The rasterization-based methods enable real-time rendering at the cost of increased memory and lower fidelity. In contrast, the ray-tracing-based methods yield superior quality but demand longer rendering time. We solve this problem by our HashPoint method combining these two strategies, leveraging rasterization for efficient point searching and sampling, and ray marching for rendering. 
Our method optimizes point searching by rasterizing points within the camera's view, organizing them in a hash table, and facilitating rapid searches. Notably, we accelerate the rendering process by adaptive sampling on the primary surface encountered by the ray. Our approach yields substantial speed-up for a range of state-of-the-art ray-tracing-based methods, maintaining equivalent or superior accuracy across synthetic and real test datasets. The code will be available at \url{https://jiahao-ma.github.io/hashpoint/}.

\end{abstract}

\vspace{-15pt}
\section{Introduction}
\label{sec:intro}
\vspace{-4pt}
Photo-realistic rendering, a significant challenge in computer vision, has seen notable advancements with NeRF~\cite{nerf} and its extensions \cite{mipnerf, mipnerf360, nerf_in_the_wild, nerf++}.
Approaches in~\cite{nerf, nerf++, nerf--, nerf, dsnerf} rely on global MLPs to reconstruct radiance fields across the entire space through ray marching. However, this method results in slow per-scene neural network fitting and extensive, often unnecessary, sampling of vast empty space, leading to prohibitive reconstruction and rendering times.
To overcome this challenge, point clouds are introduced as a straightforward representation of surfaces in space, forming point-based neural radiance fields~\cite{pointersect, pointnerf, pointslam, pbnr, npbg, npbg++, nplf, radiance_mapping, FreqPCR, neurbf}. Point clouds approximate scene geometry and encode appearance features, expediting the rendering process by sampling and aggregating features near multiple surfaces represented by the point cloud. Existing methods in this field typically adopt two rendering strategies: rasterization and ray tracing.

In the rasterization framework~\cite{npbg, npbg++, 3DGS, fwd, radiance_mapping, FreqPCR, pbnr, D3DGS, dss, surfelnerf}, points are initially projected onto the image plane, where pixel values are determined by the color of the nearest point using z-buffer mechanism. While this approach enables real-time rendering, it often exhibits visible holes stemming from the density of the point cloud. To address this limitation, some methods~\cite{npbg, npbg++, radiance_mapping, FreqPCR, neural_pc_plane_proj} leverage networks such as U-Net~\cite{unet} for hole filling by employing feature downsampling and upsampling. However, this approach struggles to generate high-fidelity details. Alternative strategies are to assign each point as a 3D shape such as an oriented disk~\cite{3DGS, D3DGS, surfels, dss, surfelnerf} or a sphere~\cite{pulsar}. Nevertheless, determining the optimal size for these shapes is a challenging task. Small shapes can lead to gaps in the rendering, while larger ones may introduce artifacts. Additionally, significant memory is often required for storage.

Ray-tracing-based point cloud rendering~\cite{pointersect, pointnerf, pointslam, trivol, snp, point2pix, nplf}, directly casts rays onto a point cloud and interpolates nearby points' features along each camera ray, addressing issues like holes found in rasterization-based methods and enabling high-fidelity novel view synthesis.
However, real-time rendering poses challenges due to the complex point cloud search process and excessive sampling of multiple surfaces.
To enhance point cloud searching efficiency, methods in~\cite{pointersect, pointnerf, pointslam, nplf} employ accelerated data structure like Uniform grid~\cite{introduction_ray_tracing}, \textit{K}-d tree~\cite{kdtree_raytracer, interative_kdtree, real_time_kdtree},
Octree~\cite{Octree, octree_pc} or bounding volume hierarchies (BVH)~\cite{bvh}. 
However, not all surface features are equally important for rendering, as illustrated in  Figure~\ref{fig:ray_tracing_process}. Typically, only surfaces closest to the camera significantly contribute to rendering, making subsequent sampling redundant. 
Other methods attempt to collect the $K$ nearest points for each ray to predict colors~\cite{nplf, pointersect}, struggling with sparse point density for extracting primary surface features.

In this work, we introduce \textbf{HashPoint}, an optimized point cloud searching approach designed to address the hole issue and accelerate the search process by adaptive sampling on primary surfaces.
The core of our method involves transforming the 3D search space into a 2D image plane for hash table lookup. 
Unlike traditional rasterization methods using the z-buffer to retain only the nearest points, we project all points onto the image plane and preserve all points within each pixel. These points are then stored in a hash table. This accelerated structure enables a swift point location near the camera ray through the hash table lookup. 
What distinguishes our approach is the adaptive searching range, determined by the distance between points and the viewpoint, rather than relying on a fixed radius or $K$ nearest points. 
After identifying the point clouds near the camera ray, these points are projected onto the ray, with each projected point serving as a potential sample point candidate. The selection of sample points is crucial, as it directly influences the feature aggregation quality on the primary surface. 
We calculate the importance of each sample point candidate based on the distance 
between the point cloud and the candidate itself. Drawing inspiration from volume rendering techniques, we retain high-importance sample point candidates, those closer to the viewpoint, becoming the definitive sample points. 
Consequently, the number of sample points for each ray varies adaptively, ranging from 0 to $n$, ensuring a dynamic sampling process.
In summary, our main contributions consist of two key techniques:
\begin{itemize}
\item 
We introduce \emph{Hashed-Point Searching} as a novel technique that accelerates the ray-tracing approach by optimizing point searching for improved efficiency.
\item 
We also propose a novel technique called \emph{Adaptive Primary Surface Sampling} to adaptively sample on the first encountered surface by the ray determined by the distance between points and the viewpoint.
\item 
We validate our approach on various benchmark datasets (synthetic, real, indoor, and outdoor), demonstrating its significant potential to accelerate the rendering process by a large margin with similar or better accuracy.
\end{itemize}
\vspace{-8pt}
\section{Related work}
\label{sec:related_work}

\vspace{-4pt}
\noindent\textbf{Point cloud rasterization.} 
Rasterization is a widely used technique for rendering point clouds. The basic concept involves projecting each point onto the image plane while ensuring that closer points occlude those that are farther away via the z-buffer mechanism. A significant challenge in point-cloud rasterization is the unwanted occurrence of gaps or holes in the rendered output. Classical methods such as visibility splatting~\cite{surfels, surface_splatting} address these gaps by substituting points with oriented disks. However, determining the optimal size and shape of these disks is complex, and they may not always completely cover the visible gaps in the image. In contrast, recent methods such as differentiable splatting, employed by \cite{3DGS, dss, surfelnerf, deepsurfels}, have made a notable improvement in rendering quality by fine-tuning the shape of these disks through optimization.
Recent advancements in the field involve the integration of rasterization with neural networks. NPBG~\cite{npbg} and NPBG++~\cite{npbg++} employ U-Net~\cite{unet} refinement to learn the rasterizing features, minimizing the disparity between the rendered images and ground truth. These approaches address the issue of holes by leveraging both feature downsampling and upsampling. Huang \emph{et al.}~\cite{radiance_mapping} introduce radiance mapping as a means to combat spatial frequency collapse. However, this technique still encounters difficulties in generating high-fidelity details in coarse regions. As a solution, FreqPCR~\cite{FreqPCR} introduces an adaptive frequency modulation module designed to capture the local texture frequency information.

\noindent\textbf{Ray tracing for point cloud.} 
This presents an alternative approach for rendering point clouds. Early works~\cite{ray_tracing_deforming_pc, splat_ray_tracing, interactive_ray_tracing_pc} developed iterative strategies to determine ray intersections with surfaces approximated from point clouds. 
Recently, methods such as NPLF~\cite{nplf} focus on embedding features at each point and aggregating them during a query, while Pointersect~\cite{pointersect} directly determines ray intersections with the inherent surface. 
PAPR~\cite{papr} employs point cloud positions to capture scene geometry, even when the initial geometry substantially differs from the target geometry. However, this method mainly focuses on the nearest points around the ray, potentially missing the features of the primary surface, particularly when dealing with sparse point clouds.
On the other hand, Point-SLAM~\cite{pointslam} and Point-NeRF~\cite{pointnerf}  uniformly sample across multiple surfaces to produce high-quality renderings. Despite avoiding unnecessary sampling of vast empty space, such processes of extensive feature aggregation and MLP prediction on subsequent surfaces remain time-consuming. Thus, there is a need for selective sampling on correct surfaces to improve both accuracy and efficiency in point cloud rendering.

\noindent\textbf{Efficiency for neural radiance field.} Recent works~\cite{mixnerf, f2nerf, instant-ngp, neus2} improve the memory and optimization efficiency via hash encoding. 
Some other works~\cite{r2l, donerf, terminerf, autoint, adanerf} are designed to improve the sampling efficiency of NeRF while maintaining a compact memory footprint. 
DONeRF~\cite{donerf} improves sampling efficiency with a depth-trained oracle network. TermiNeRF~\cite{terminerf} uses a density-based sampling network from a pre-trained NeRF. AutoInt~\cite{autoint} predicts segment lengths along rays for sampling. 
AdaNeRF~\cite{adanerf} employs a dual-network architecture with incremental sparsity for fewer samples of higher quality. 
In general, these methods utilize pre-trained modules and complex optimization for sample point distribution along rays.
In addition to neural network-assisted sampling, some other approaches~\cite{diner, ddnerf, mipnerf_rgbd, enerf} adopt multi-view stereo (MVS) technique~\cite{mvs, mvsnet} or depth sensor to obtain depth information\cite{pointslam}, subsequently sampling only on the surfaces of scenes, closely resembling ray-tracing based point cloud rendering. Yet, they rely on dense depth map input and tend to encounter inefficiencies from oversampling on multiple surfaces during rendering. This paper presents our pioneer concept of sampling on the primary surface, further optimizing the use of the explicit geometric representation.

\vspace{-6pt}
\section{Method}
\label{sec:Method}

\begin{figure*}[ht]
\centering
\includegraphics[width=0.9\linewidth]{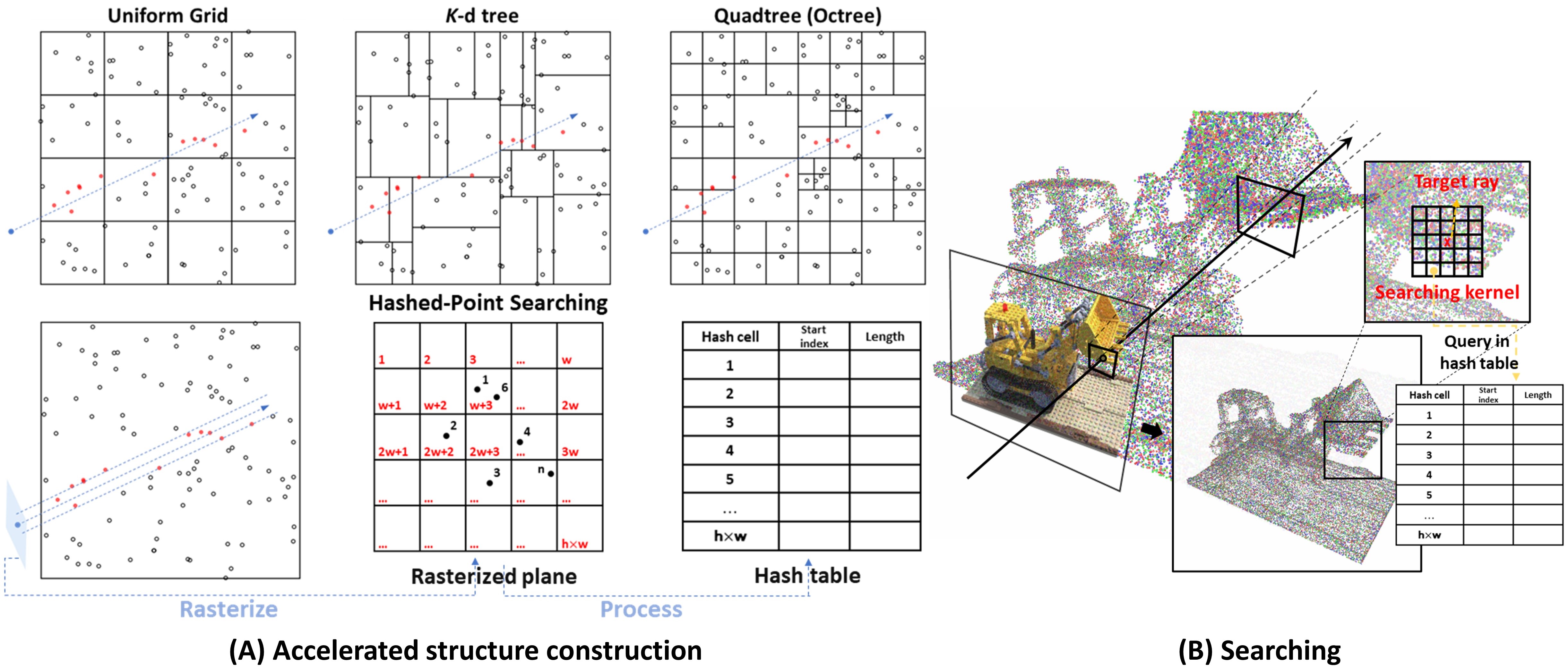}
\caption{Efficient point clouds searching for ray tracing. 
\textbf{(A) Top row} shows traditional point cloud search strategies: Uniform Grid, \textit{K}-D tree, and Octree visualized in 2D for clarity. 
\textbf{(A) Bottom row} shows \emph{Hashed-Point Searching}, our method of rasterizing the point cloud onto an image plane and then reorganizing the point cloud list to optimize queries, resulting in the construction of a hash table. Each key (as pixel index) in the table maps to the start index in the point list and the count of points within that pixel with $\mathcal{O}(1)$ complexity. 
\textbf{(B)} The final selection depicts the search mechanism: using a magnified search kernel, the neighbor points of a target ray (black arrow) are swiftly identified through the hash table and assessed based on their distance to the ray. 
}
\label{fig:point_ray_searching}
\vspace{-8pt}
\end{figure*}

Following preliminaries in Section~\ref{Method:Preliminary}, we detail our HashPoint method with our point searching technique in Section~\ref{Method:point_searching}, and our surface sampling approach in Section~\ref{Method:point_selection}, and then integrate these with existing methods in Section~\ref{Method:Integration}. Comparative analysis is explained in Section~\ref{sec:discussion}.

\subsection{Preliminaries of neural rendering}
\label{Method:Preliminary}
Point cloud rendering based on ray tracing typically employs two strategies to ascertain the color of each ray:

\noindent\textbf{Image-based rendering.} Given a point cloud denoted as $\mathcal{P}=$ $\left\{\left(x^{pc}_i, c^{pc}_i, f_i \right)\right\}_{i=1 \ldots n}$, where $x^{pc}_i \in \mathbb{R}^3$ represents position, $c^{pc}_i \in \mathbb{R}^3$ is RGB color and $f_i$ depicts high-dimensional appearance feature. 
Several methods~\cite{pointersect, nplf} predict the blending weight $w_i(\mathbf{r}, \mathcal{P})$ for camera ray $\mathbf{r}$ utilizing the appearance feature, while others, like \cite{nplf, papr}, use the geometric relationship with the ray. The final color of a camera ray $c(\mathbf{r})$ can be computed by \textit{K} nearest points, represented as \textit{K-NP} in Figure~\ref{fig:sampling}A: 
\vspace{-6pt}
\begin{equation}
c(\mathbf{r})=\sum_{i=1}^K w_i(\mathbf{r}, \mathcal{P}) c^{pc}_i.
\vspace{-4pt}
\end{equation}

Although methods \cite{pointersect, nplf, papr} exhibit some variations, fundamentally, they all harness the relationship between rays and proximate points to determine color.

\noindent\textbf{Volume rendering.} Unlike predicting colors from neighboring points of a ray, some methods \cite{pointnerf, pointslam} uniformly sample on surfaces, and aggregate point features $f_i$ to sampled points, as shown in Figure~\ref{fig:sampling}B. These methods subsequently employ physically-based volume rendering to compute the color of individual pixels. Specifically, the radiance of a pixel is derived by traversing a ray through it, sampling N sample points at $\left\{x^{sp}_j \mid j=1, \ldots, N\right\}$ along the ray, and accumulating radiance using volume density, as:
\vspace{-6pt}
\begin{equation}
\begin{aligned}
c & =\sum_{j=1}^N \tau_j\left(1-\exp \left(-\sigma_j \Delta_t\right)\right) c^{sp}_j, \\
\tau_j & =\exp (-\sum_{k=1}^{j-1} \sigma_k \Delta_k) .
\end{aligned}
\vspace{-4pt}
\end{equation}
Here, $\tau$ represents volume transmittance; $\sigma_j$ and $c^{sp}_j$ correspond to the density and color for each sample point $j$ at $x^{sp}_j$. $\Delta_t$ is the distance between adjacent sample points.

\subsection{Hashed-point searching}
\label{Method:point_searching}

\noindent\textbf{Accelerated structure construction.} 
In contrast to traditional approaches that perform point cloud searching in 3D space, we innovate by shifting the search from 3D space to a 2D image plane for hash table lookup. As illustrated in Figure~\ref{fig:point_ray_searching}, our approach diverges from classic rasterization methods that employ the z-buffer to retain the nearest points for each pixel. Instead, we project all points onto the image plane and preserve all points within each pixel, storing them in a hash table for efficient retrieval.
Our HashPoint approach to reorganized point cloud is inspired by the Morton code \cite{morton_code} (often referred to as Z-order), which provides a linear ordering of multi-dimensional data for efficient retrieval and storage. Specifically, we arrange the points falling on the same pixel in a Z-order manner to ensure they are stored as closely as possible in sequence. Therefore, we adjust the order of points in the point cloud list. The positional adjustment of each point is achieved through atomic operations in CUDA~\cite{cuda}, with a time complexity of only $\mathcal{O}(1)$ for high efficiency. Following these adjustments, we use a hash table to store the position of each point in the point list. In the hash table, the key is the index of the pixel, and the value includes the position of the first point stored for the current pixel in the point list, along with the number of points that fall on that pixel.

\noindent\textbf{Searching.} Inspired by \cite{mipnerf, trimiprf, zipnerf} each pixel emits a cone for point cloud searching and feature aggregation, which is more in line with the principles of imaging. The use of a fixed searching radius may include features that do not belong to the pixel, introducing noise. For a detailed comparison, please refer to our Supplementary. In this work, we propose an adaptive searching radius that is proportional to the distance of the sampling point relative to the ray origin. 

\begin{figure}[t!]
\centering
\includegraphics[width=0.9\linewidth]{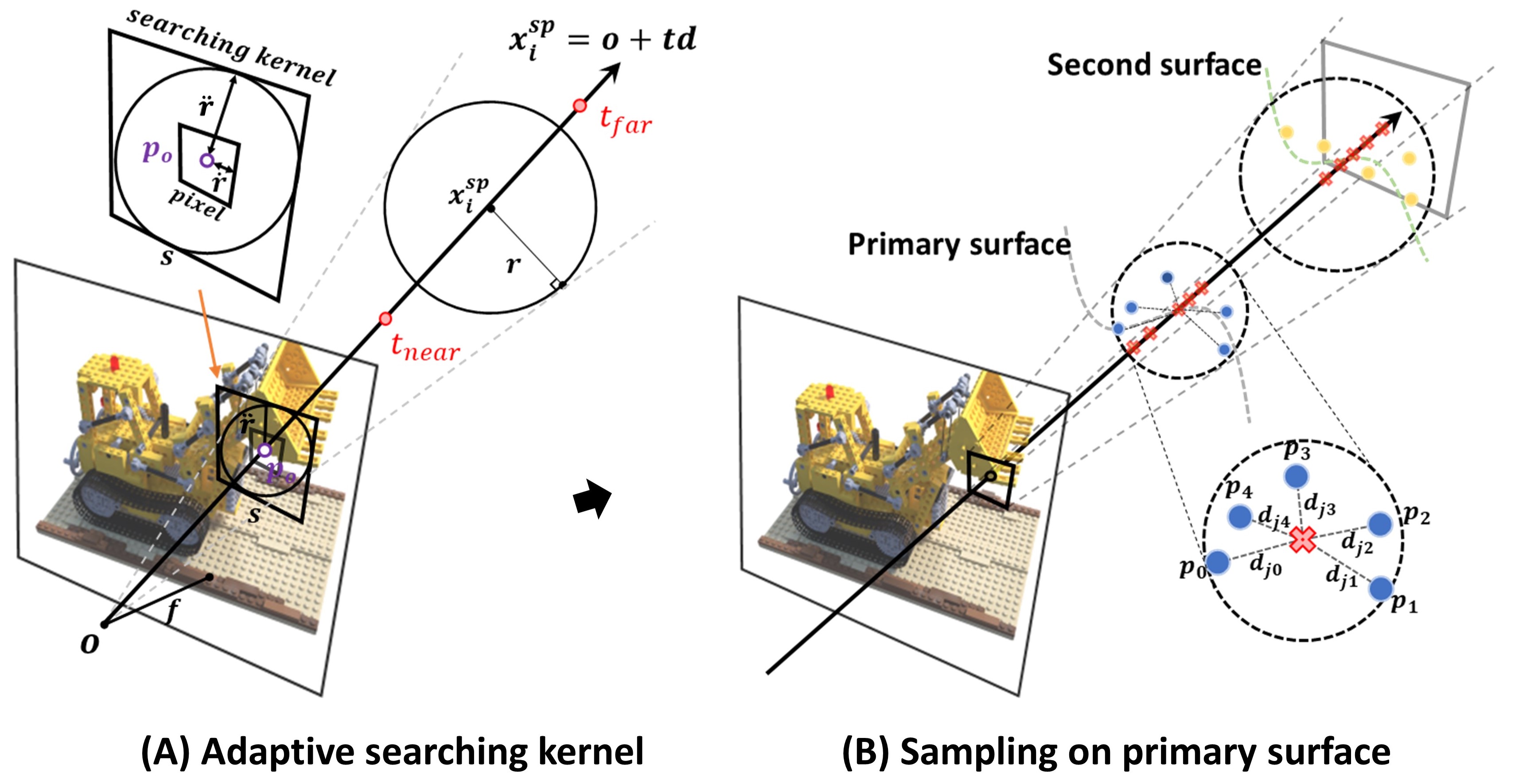}
\caption{Illustration of \emph{Adaptive Primary Surface Sampling}. 
\textbf{(A)} The diagram depicts the generation of the searching kernel on the image plane. 
\textbf{(B)} We project adjacent points to the ray as sample point candidates (red crosses). Each candidate's importance, determined by its distance distribution to points within its radius, influences its preservation for final feature aggregation.}
\vspace{-9pt}
\label{fig:adaptive_sampling}
\end{figure}

Following~\cite{trimiprf}, we represent the pixel as a circle on the image plane, as an approximation to the area of the pixel. As shown in Figure~\ref{fig:adaptive_sampling}A, the radius of the disc can be calculated by  $\dot{r}=\sqrt{\Delta x \cdot \Delta y / \pi}$, where $\Delta x$ and $\Delta_y$ are the width and height of the pixel in world coordinates, derived from the calibrated camera parameters. For each pixel, a cone emits from the ray origin $\mathbf{o}$ (optical center of the camera) along the ray direction $\mathbf{d}$, passing through the pixel center $\mathbf{p_o}$.  Due to the sparsity of the point cloud, there might be no points falling within the ray cone, causing holes in the rendering. To mitigate this, we modify the ray cone for broader coverage, shifting from a disc with radius $\dot{r}$ for one pixel to a larger disc of radius $\ddot{r}$ for the searching kernel. We deduce the magnitude of the searching kernel $s$ by $s = 2 \cdot \left\lceil \frac{\ddot{r}}{\dot{r}} \right\rceil + 1$. Any sample point that lies on the ray can be derived by $\mathbf{x^{sp}_j} = \mathbf{o} + t \mathbf{d}$, and the adaptive radius $r$ of sample point can be computed by, 

\vspace{-7pt}

\begin{figure}[h]
\vspace{-8pt}
\resizebox{0.45\textwidth}{!}{
\begin{minipage}{0.5\textwidth}
\begin{equation}
\begin{aligned}
    r = \frac{\|\mathbf{x^{sp}_j}-\mathbf{o}\|_2 \cdot f \ddot{r}}{\|\mathbf{p_o} - \mathbf{o}\|_2 \cdot \sqrt{\left(\sqrt{\|\mathbf{p_o} - \mathbf{o}\|_2^2-f^2}- \ddot{r}\right)^2+f^2}.} 
\label{eq:radius}
\end{aligned}
\end{equation}
\end{minipage}
}
\vspace{-10pt}
\end{figure}
Here, $f$ is the focal length. For derivation, please refer to the Supplementary.  
During the rendering process, we sample between $t_{near}$ and $t_{far}$. As $t$ increases, the radius of the sample point also increases. As the minimum radius, $r_{min}$, is at the $t_{near}$ position and matches with our input hyperparameter radius, we modify $s$ accordingly.  In the target ray's search kernel, we traverse each pixel and assess the number of associated points per pixel to determine if the search is needed. If required, we look up the point list for further judgment. The query complexity remains at $\mathcal{O}(1)$. 

\subsection{Adaptive primary surface sampling} 
\label{Method:point_selection}
Another core of our proposed method lies in the adaptive aggregation of features from the primary surface, guided by the distribution of the nearby point cloud. As shown in Figure~\ref{fig:adaptive_sampling}B, neighboring points are projected onto the camera ray, forming a set of sample point candidates denoted as $x^{sp}_j$ where $j$ represents the index of the sample point. The geometric distribution of these sample points is determined by calculating the average distance between each candidate and its \textit{K} neighboring points. The average distance $d_j$ can be computed by
\vspace{-6pt}
\begin{equation}
\begin{aligned}
    d_j = \frac{1}{K} \sum_{i=1}^{K}\left\|x_j^{sp}-x_i^{p c}\right\|_2 .
\end{aligned}
\vspace{-4pt}
\end{equation}
While this distance between the sample point and the surface presented by the point cloud adheres to the principles of the Unsigned Distance Function (UDF)~\cite{neuraludf}, it deviates from a strict UDF definition. Therefore, we term it as ``pseudo-UDF''.
Drawing inspiration from volume rendering \cite{nerf} and its extension \cite{neus, neuraludf}, we adhere to two rules: 
\begin{enumerate}
\item \textbf{Unbiased intersection}. A reduced pseudo-UDF value indicates that the candidate is proximal to the surface, thus deserving a higher weighting.
\item \textbf{Occlusion-aware}. When candidates exhibit identical pseudo-UDF values, the one positioned closer to the viewport is assigned a greater weight to the final color output.
\end{enumerate}
To follow the first principle, the distance $d_j$ is transformed into a confidence $a_j$ by $\alpha_j = \gamma\exp \left(-\frac{d_j^2}{\beta^2}\right),$
where $\beta$ is the hyperparameter which depends on the density of the point cloud and $\gamma$ is to control the range of sampling. The proximity of the sample point to the point cloud, lesser $d_j$, necessitates a reduced $\beta$ to ensure distinct differentiation in the confidence $a_j$ cross sample points. To make the distribution of sample point occlusion aware, based on volume rendering, we define the weight function by
\vspace{-6pt}
\begin{equation}
w_j =\alpha_j \prod_{k=1}^{j-1}\left(1-\alpha_k\right).
\label{eq:weight}
\vspace{-4pt}
\end{equation}
We retain the sample point candidates with $w_j$ larger than zero, which are located near the primary surface.

\subsection{Integration with existing methods}
\label{Method:Integration}

The proposed method is adaptable to most ray-tracing point cloud rendering techniques. After sampling on the primary surface and identifying surface point cloud, we can use MLP, as in \cite{pointslam, pointnerf}, to predict density and color for sample points. The color of the pixel is then rendered by volume rendering. Additionally, we can integrate approaches from \cite{pointersect, nplf, papr} to estimate the ray's color based on point features. Please refer to Section \ref{sec:experiment} for more comparison.

\subsection{Comparative analysis}
\label{sec:discussion}

\noindent \textbf{Our hashed-point searching \textit{vs.} traditional point cloud searching.} 
Efficient retrieving neighboring points for a camera ray is non-trivial. A straightforward method is to compare all rays with all points, brute force searching, leading to high computation costs. Common approaches employ space partitioning to speed up the search. Below, taking an example with $n$ points and $m$ rays with an average of $q$ points falling in the radius $\delta$ of each ray, we introduce the search processes and the corresponding complexity of these algorithms.

\begin{itemize}
    \item \textbf{\textit{K}-d tree} \cite{original_kdtree}: a space-partitioning data structure for organizing points in a k-dimensional space. To locate points near a ray using this structure, two common strategies exist: (1) Sample uniformly along the ray followed by radius searches at each sample point, and (2) leverage sub-tree bounding boxes in the \textit{K}-d tree and apply axis-aligned bounding box (AABB \cite{geometric_tools}) for intersections, then compare points in intersecting sub-trees. Building the \textit{K}-d tree needs to iterate $n$ points with the complexity of $\mathcal{O}(n)$, and in the case of searching points for $m$ rays, the complexity of finding all intersecting sub-trees and retrieving $q$ points within the cylinder of rays is $\mathcal{O}(mlog(n) + mq)$.

    \item \textbf{Octree} \cite{Octree}: a tree data structure used to partition a three-dimensional space by recursively subdividing it into eight octants. This strategy for finding points near a ray is similar to the \textit{K}-d tree mentioned above, and the complexity is the same - constructing an Octree requires $\mathcal{O}(n)$, while searching requires $\mathcal{O}(mlog(n) + mq)$.
    
    \item \textbf{Uniform grid} \cite{introduction_ray_tracing}: a data structure that divides space into regular grids and allocates data to each grid cell. Building this data structure requires traversing all points with complexity $\mathcal{O}(n)$. For searches, using algorithms like 3DDA \cite{voxel_traversal} or AABB to find intersecting voxel with rays, the complexity is $\mathcal{O}(mg + mq)$, where $g$ is the number of grid cells per dimension.

    \item \textbf{Hashed-point searching}: In terms of design, prior methods involve the ray actively seeking point cloud in 3D space. In contrast, our method allows the point cloud to seek rays. To accelerate queries, we construct an acceleration structure in the form of a hash table by iterating through all points, followed by searching via hash table lookup. The complexity of this process is $\mathcal{O}(n + mq)$.
\end{itemize}

\begin{figure}[t!]
\centering
\includegraphics[width=1.0\linewidth]{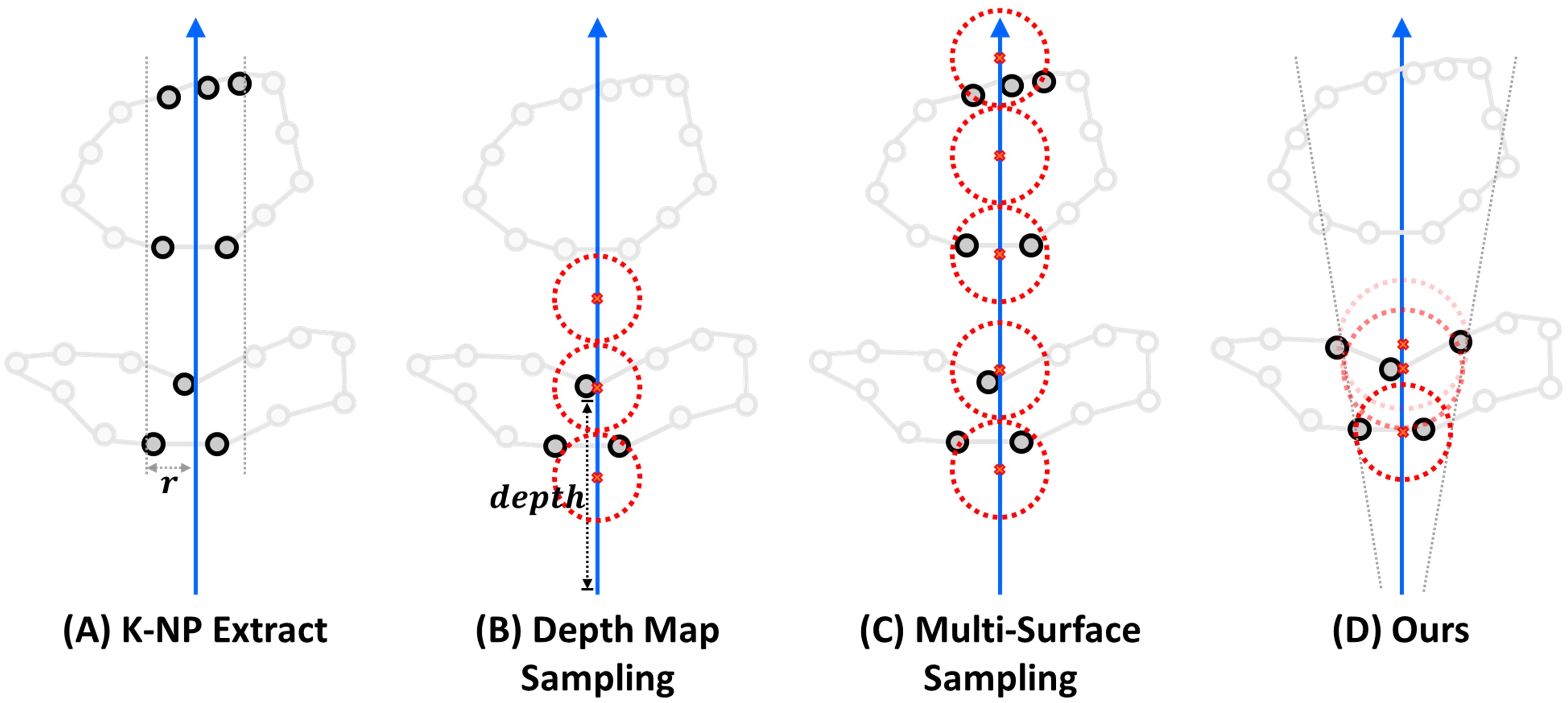}
\caption{Comparison of different point cloud selection strategies. 
\textbf{(A)} \textit{K-NP Extract}: extracting \textit{K} nearest point features per ray. 
\textbf{(B)} \textit{Depth Map Sampling}: sampling based on dense depth maps. 
\textbf{(C)} \textit{Multi-Surface Sampling}: uniform sampling over multiple surfaces. 
\textbf{(D)} \textit{Our method}: adaptive sampling on the primary surface. }
\label{fig:sampling}
\vspace{-13pt}
\end{figure}

\noindent \textbf{Our adaptive primary surface sampling \textit{vs.} existing point cloud selection strategies.} 
The selection of the number and location of points significantly determines the quality and efficiency of rendering. As shown in Figure~\ref{fig:sampling}, we compare various methods of point cloud selection as follows:
\begin{itemize}
    \item ``\textit{K-NP Extract}''~\cite{papr, pointersect, nplf} refers to extracting the nearest \textit{K} point features around a ray to predict its color. As shown in Figure~\ref{fig:vis_comparison_with_nplf}, due to the density of point cloud, the \textit{K} nearest points often don't necessarily fall on the primary surface, leading to feature noise. 

    \item ``\textit{Depth Map Sampling}''~\cite{diner, pointslam} refers to using sensor depth or predicted depth via MVS~\cite{mvs, mvsnet} to assist sampling. Although this method is highly efficient, it relies on dense depth maps and also faces the issue of redundant sampling on multiple surfaces (especially in the case of relying on nearby views with depth input). 

    \item ``\textit{Multi-Surface Sampling}''~\cite{pointnerf} involves finding all surfaces intersecting with the ray and then uniformly sampling on them. The method provides the best result at the high cost of computation. 

    \item \textit{"Adaptive Primary Surface Sampling"} gives a good balance between precision and efficiency by sampling on the primary surface, as explained in Sections~\ref{Method:point_selection},~\ref{exp:evaluation} and~\ref{exp:ablation_study}.
\end{itemize}

\noindent \textbf{Ours \textit{vs.} 3D Gaussian splatting~\cite{3DGS}.} While our method shares similarities with 3D Gaussian splatting in utilizing rasterization for point searching, a key distinction lies in our rendering approach. While Gaussians have shape, they are often small, requiring more points for ray rendering. In contrast, our method \textit{interpolates} ray color from nearby points, requiring fewer points. For instance, in experiments with the same Lego scene (NeRF-Synthesis Dataset), our method uses only 35Mb of storage, compared to Gaussian splatting's 200Mb. The proposed method takes advantage of both rasterization and ray tracing, complementing each other. The combination of the two methods further enhances the approach to promising neural rendering. 

\vspace{-6pt}
\section{Experiments}
\label{sec:experiment}






\subsection{Baselines and integration}
This section outlines four key baselines in ray-tracing point cloud rendering, focusing on searching and sampling, as detailed in Table~\ref{tab:baselines}. We compare the corresponding components with the proposed enhancement as follows: 

\begin{itemize}
    \item \textit{Point-NeRF} \cite{pointnerf}: Searches via uniform grid (\textit{UG}) and then samples across multiple surface (\textit{MS}). We replace the \textit{UG} and \textit{MS} with our HashPoint (\textit{HS}) and primary surface sampling (\textit{PS}) and benchmark as shown in Table \ref{tab:comparison_pointnerf}. \textit{PS} alone impedes the point optimization due to limited gradient propagation. Thus, \textit{MS} is initially used for 10$K$ iterations for geometry optimization, then switched to \textit{PS} for boosting efficiency. The transition is controlled by adjusting the parameter $\gamma$. For more details, please refer to the supplementary material. 
    
    \item \textit{Point-SLAM} \cite{pointslam}: Employs single-surface (\textit{SS}) sampling, different from Point-NeRF's \textit{MS} approach. While Point-SLAM's depth-guided sampling is efficient, it relies on dense depth input, not always available during rendering. Our method's performance as shown in Table~\ref{tab:comparison_pointslam} is compared with both of Point-SLAM's sampling strategies: depth-guided and uniform multi-surface.
    
    \item \textit{NPLF} \cite{nplf}: Starts with downsampling points, searches nearby points using brute force (\textit{BF}),  and renders on \textit{K} nearest points (\textit{K-NP}). Our method maintains high fidelity by selecting \textit{K} (\textit{K} = 8) nearest points from the primary surface without downsampling as shown in Table \ref{tab:comparison_nplf}.
    
    \item \textit{Pointersect} \cite{pointersect}: Searches using a \textit{UG}, retaining \textit{K} (\textit{K} = 40) nearest points for rendering. We improve this by selecting \textit{K} (\textit{K} = 6) nearest points from primary surfaces, gaining higher speed and quality as shown in Table \ref{tab:comparison_pointersect}.
    
\end{itemize}

We evaluate our search and selection modules, along with overall performance, in Sections \ref{exp:evaluation} and \ref{exp:ablation_study}. The evaluation metrics include Peak Signal-to-Noise Ratio (PSNR), Structural Similarity Index Measure (SSIM), Learned Perceptual Image Patch Similarity (LPIPS), and frames per second (FPS). Our method, integrated with these baselines, is tested on Synthetic-NeRF \cite{nerf}, Waymo \cite{waymo}, Replica \cite{replica}, and ShapeNet \cite{shapenet} datasets. We adopt their training approach, with a key difference: in Point-NeRF, we refine the geometry, unlike other experiment baselines. This refinement impacts accuracy; for more details, see Section~\ref{ablation:geometry_refine}.

\begin{table}
    \centering
    \resizebox{0.43\textwidth}{!}{
    \begin{tabular}{ccc}
    \toprule
        \textbf{Method} & \textbf{Searching} & \textbf{Selection}\\
    \midrule
        Point-NeRF~\cite{pointnerf} & Uniform grid & Multiple surfaces \\
        NPLF~\cite{nplf} & Brute force & \textit{K} nearest points\\
        Pointersect~\cite{pointersect} & Uniform grid & \textit{K} nearest points\\
        Point-SLAM~\cite{pointslam} & Depth map sampling & Single surface\\
    \bottomrule
    \end{tabular}
    }
    \vspace{-6pt}
    \caption{Comparison of key components in various baselines.}
    \label{tab:baselines}
\vspace{-12pt}
\end{table}

\begin{table*}
    \centering
    \resizebox{1.0\textwidth}{!}{
    \begin{tabular}{lccccccc}
        \toprule
        & Radiance-based & \multicolumn{4}{c}{Point-based (Rasterization)} & \multicolumn{2}{c}{Point-based (Ray tracing)} \\
        \cmidrule(lr){2-2} \cmidrule(lr){3-6} \cmidrule(lr){7-8}
        & NeRF \cite{nerf} & NPBG~\cite{npbg} & NPBG++~\cite{npbg++} & Huang \textit{et al.}~\cite{radiance_mapping} & FreqPCR~\cite{FreqPCR} & Point-NeRF~\cite{pointnerf} & Point-NeRF + Ours\\
        \midrule
        PSNR $\uparrow$  & 31.01 & 24.56  & 28.12 & 28.96 & 31.24 & \textbf{33.31} & 33.22\\
        SSIM $\uparrow$  & 0.947 & 0.923  & 0.928 & 0.932 & 0.950 & \textbf{0.978} & 0.961 \\
        LPIPS$\downarrow$& 0.081 & 0.109  & 0.076 & 0.061 & 0.049 & \textbf{0.049} & 0.055\\
        FPS  $\uparrow$  & 0.05 & 33.64   & 35.21  & \textbf{39.67}  & 39.56  & 0.12 & 9.60($\times$80 speed up)\\
        \bottomrule
    \end{tabular}}
    \vspace{-6pt}
    \caption{Comparison of our method integrated with Point-NERF with a radiance-based model~\cite{nerf}, rasterization-based models~\cite{npbg,npbg++,radiance_mapping,FreqPCR} and a ray-tracing-based model~\cite{pointnerf} on the Synthetic-NeRF dataset. 
    }
    \label{tab:comparison_pointnerf}
\vspace{-8pt}
\end{table*}

\vspace{-4pt}
\subsection{Evaluation}
\label{exp:evaluation}

\begin{figure}[t]
\centering
\vspace{-4pt}
\includegraphics[width=1\linewidth]{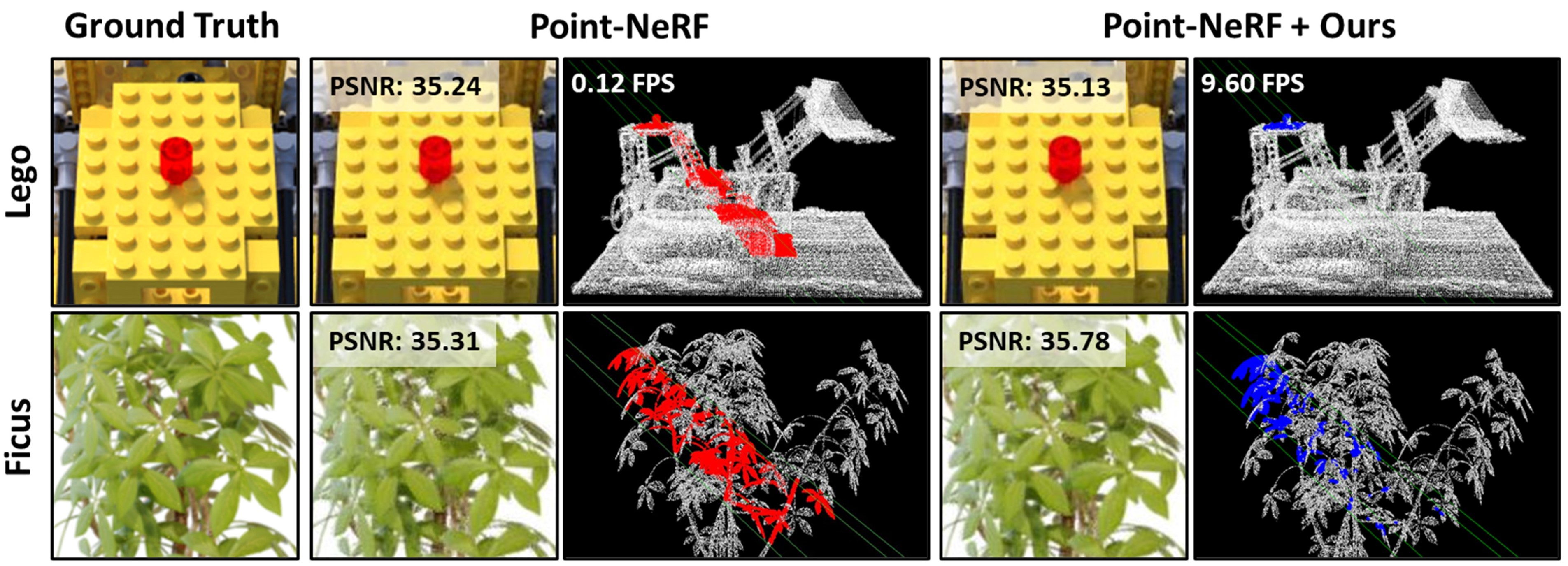} 
\caption{
Results on the NeRF-Synthesis \cite{nerf} dataset shows that our primary surface sampling (\textcolor{blue}{blue points}) is more efficient than Point-NeRF's sampling (\textcolor{red}{red points}) while preserving accuracy.
}
\label{fig:vis_comparison_with_pointnerf}
\vspace{-8pt}
\end{figure}

\noindent \textbf{Evaluation on Synthetic-NeRF \cite{nerf}.} We incorporate the proposed method into Point-NeRF \cite{pointnerf} and compare with both NeRF \cite{nerf} and various point-based approaches~\cite{npbg, npbg++, radiance_mapping, FreqPCR, pointnerf}. As shown in Table~\ref{tab:comparison_pointnerf}, this integration achieves an \textbf{80}-fold speedup while ranking as the second-highest performance among point-based methods. The results of Figure~\ref{fig:vis_comparison_with_pointnerf} qualitatively show that solely sampling on the primary (single) surface can produce multi-surface effects, presenting the right balance between performance and speed for ray-tracing-based methods.

\noindent \textbf{Evaluation on Replica \cite{replica}.} We also compare the integration of Point-SLAM \cite{pointslam} with voxel-based method \cite{niceslam}  and the original \cite{pointslam} on the Replica \cite{replica} indoor dataset, solely focusing on rendering rather than tracking and mapping. As outlined in Table~\ref{tab:comparison_pointslam}, our approach outperforms Point-SLAM in both PSNR and SSIM with parity in LPIPS. In terms of speed, we analyzed Point-SLAM's depth-guided and uniform multi-surface sampling (\textit{US}), particularly for depth-unknown scenarios. 
In contrast to Point-SLAM's fixed $\hat{n}$ point collection ($\hat{n} = 5$ ), our method dynamically gathers 1 to \textit{n} points based on nearby point distributions ($\textit{n} = 4$). The efficiency of our approach is notably superior - 1.8 times faster than depth-guided sampling and 11.5 times faster than \textit{US} sampling.

\begin{figure}[t]
\centering
\includegraphics[width=0.96\linewidth]{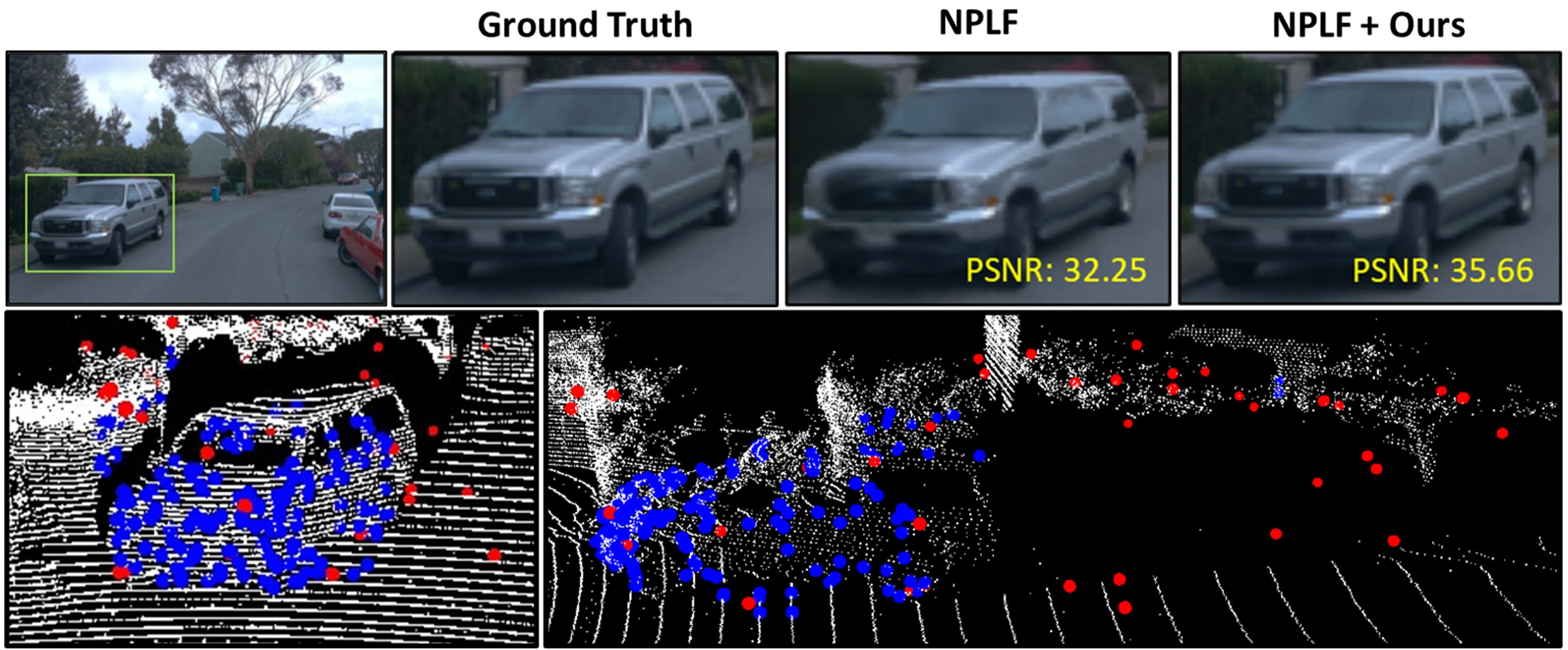} 
\caption{
Comparison on Waymo~\cite{waymo} dataset shows how our primary surface sampling (\textcolor{blue}{blue points}) more accurately samples the car body than the \textit{K} nearest point selection of NPLF (\textcolor{red}{red points}).
}
\label{fig:vis_comparison_with_nplf}
\vspace{-10pt}
\end{figure}

\noindent \textbf{Evaluation on Waymo \cite{waymo}.} In the evaluation of the Waymo dataset, we substitute the \textit{BF} and \textit{K-NP} with our methods while comparing radiance-based methods~\cite{nerf, dsnerf} and the original~\cite{nplf}. Figure~\ref{fig:vis_comparison_with_nplf} illustrates that our method accurately samples object surfaces, outperforming \textit{K-NP}. Table~\ref{tab:comparison_nplf} shows our method's superior accuracy over NPLF and its enhanced efficiency compared to \textit{BF}.

\noindent \textbf{Evaluation on ShapeNet \cite{shapenet}.} To evaluate on the ShapeNet dataset without per-scene optimization, we adopt the training setting of Pointersect - training on 48 training meshes from the sketchfab \cite{pugeonet} dataset and testing on 30 meshes in ShapeNet. We streamline the architecture of Pointersect by reducing the input of neighbor points (\textit{K}: \textit{40} to \textit{6}), significantly enhancing the efficiency of inference. The results presented in Table~\ref{tab:comparison_pointersect} indicate that using only the \textit{K} nearest points of the primary surface produces robust outcomes and accelerates the process.

\begin{table}
    \centering
    \resizebox{0.48\textwidth}{!}{
    \begin{tabular}{lccc}
    \toprule
    &  \multicolumn{1}{c}{Voxel-based} &  \multicolumn{2}{c}{Point-based (Ray tracing)} \\
    \cmidrule(lr){2-2} \cmidrule(lr){3-4}

    & NICE-SLAM~\cite{niceslam} & Point-SLAM~\cite{pointslam} & Point-SLAM + Ours\\
    \midrule
    PSNR $\uparrow$  & 24.42 & 35.17 & \textbf{35.43}\\
    SSIM $\uparrow$  & 0.809 & 0.975 & \textbf{0.983}\\
    LPIPS$\downarrow$& 0.233 & \textbf{0.124} & 0.126\\
    FPS  $\uparrow$  & 0.43 & 0.95($Depth$) $\vert$ 0.15($US$) & \textbf{1.72} ($\times$1.8 $\vert$ $\times$11.5 speed up)\\
    \bottomrule
    \end{tabular}}
    \vspace{-6pt}
    \caption{Comparison of our method integrated with Point-SLAM \cite{pointslam} with NICE-SLAM \cite{niceslam} and Point-SLAM on Replica dataset \cite{replica}. $Depth$ and $US$ denote depth-guided sampling on a single surface and uniformly sampling across multiple surfaces separately. The speed evaluates the performance of rendering instead of mapping and tracking.}
    \label{tab:comparison_pointslam}
\vspace{-8pt}
\end{table}

\begin{table}
    \centering
    \resizebox{0.48\textwidth}{!}{
    \begin{tabular}{lcccc}
    \toprule
    &  \multicolumn{2}{c}{Radiance-based} &  \multicolumn{2}{c}{Point-based (Ray tracing)} \\
    \cmidrule(lr){2-3} \cmidrule(lr){4-5}
    & NeRF \cite{nerf} & DS-NeRF\cite{dsnerf} & NPLF \cite{nplf} & NPLF + Ours\\
    \midrule
    PSNR $\uparrow$  & 22.47 & 26.15 & 29.96 & \textbf{30.57}\\
    SSIM $\uparrow$  & 0.700 & 0.772 & 0.868 & \textbf{0.912}\\
    LPIPS$\downarrow$& 0.389 & 0.310 & 0.119 & \textbf{0.105}\\
    FPS  $\uparrow$  & 0.11 & 0.11 & 0.33 & \textbf{1.98} ($\times$ 6 speed up)  \\
    \bottomrule
    \end{tabular}}
    \vspace{-6pt}
    \caption{Comparison of our method integrated with NPLF \cite{nplf} with two radiance-based model~\cite{nerf,dsnerf}, and a ray-tracing-based model NPLF on the Waymo Open dataset.
    }
    \label{tab:comparison_nplf}
\vspace{-8pt}
\end{table}

\begin{figure}[t]
\centering
\includegraphics[width=0.96\linewidth]{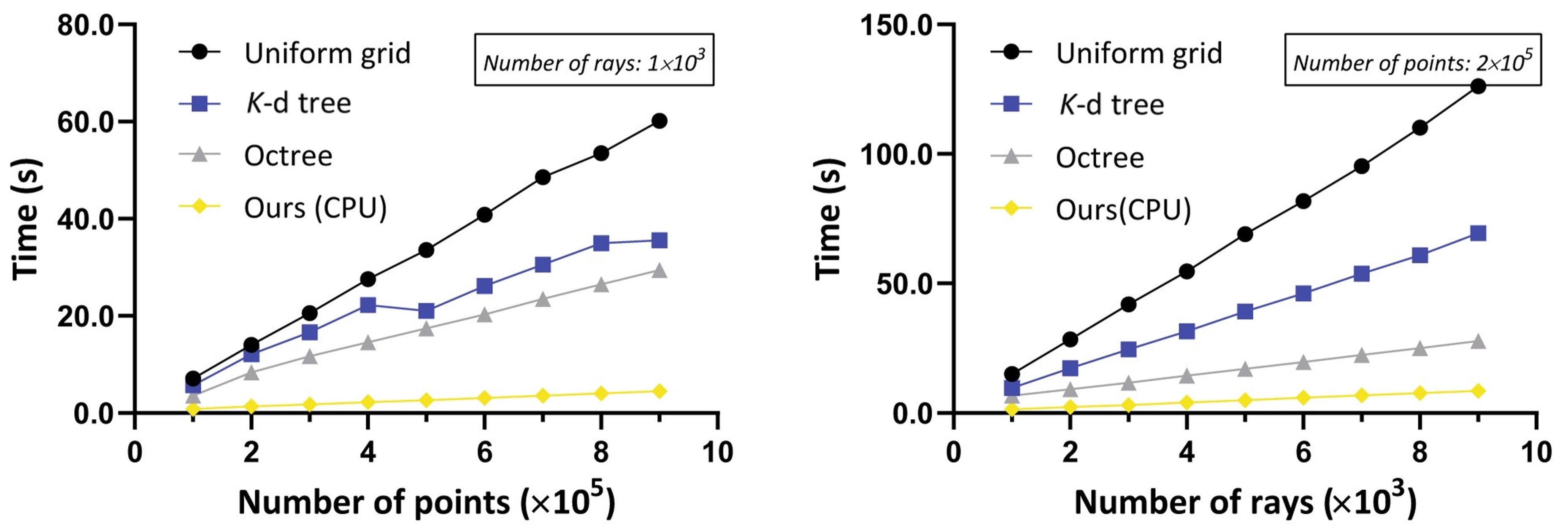}
\caption{
Comparative search performance for neighbor points searching for ray casting: ours \textit{vs.} uniform grid \cite{introduction_ray_tracing}, \textit{K}-d tree \cite{original_kdtree} and Octree \cite{Octree}. 
\textbf{Left}: search times for various numbers of points with a fixed number of rays. \textbf{Right}: search time for a fixed number of points with increasing numbers of rays.
}
\vspace{-6pt}
\label{fig:point_ray_searching_output}
\end{figure}

\noindent \textbf{Comparison with traditional point cloud search for ray casting.} 
In Figure~\ref{fig:point_ray_searching_output}, our method outperforms traditional methods (Uniform grid, \textit{K}-d tree, and Octree) in point cloud search efficiency for ray casting. All comparisons are conducted on a CPU (\textit{Intel(R) Core(TM) i9-12900K}) without GPU acceleration. The left graph shows our method's superior performance with increasing point numbers, while the right graph demonstrates consistent efficiency with rising ray counts. Notably, all methods retrieve the same number of points, underscoring that differences in performance result from search efficiency.
Our proposed method running on a \textit{NVIDIA RTX 4090} GPU for 1 million points only takes 4 ms to complete searching (0.5 ms) and sampling (3.5 ms) proving its high efficiency of processing large-scale point cloud data for ray casting.

\begin{table}
    \centering
    \resizebox{0.48\textwidth}{!}{
    \begin{tabular}{lccc}
    \toprule
    & \multicolumn{3}{c}{Point-based}  \\
    \cmidrule(lr){2-4}
    &  \multicolumn{1}{c}{Rasterization} &  \multicolumn{2}{c}{Ray tracing} \\
    \cmidrule(lr){2-2} \cmidrule(lr){3-4}
    & NPBG++~\cite{npbg++} & Pointersect ~\cite{pointersect} & Pointersect + Ours\\
    \midrule
    PSNR $\uparrow$  & 19.3 $\pm$ 4.0 & 28.0 $\pm$ 6.4 & \textbf{29.5 $\pm$ 5.6}\\
    SSIM $\uparrow$  & 0.8 $\pm$ 0.1& 1.0 $\pm$ 0.0 &  \textbf{1.0 $\pm$ 0.0}\\
    LPIPS$\downarrow$& 0.18 $\pm$ 0.08& 0.04 $\pm$ 0.04& \textbf{0.02 $\pm$ 0.01}\\
    FPS  $\uparrow$  & \textbf{33.40} & 1.25 &  10.12 ($\times$8 speed up) \\
    \bottomrule
    \end{tabular}}
    \vspace{-6pt}
    \caption{ Comparison on the ShapeNet \cite{shapenet} dataset shows that the integration of our approach with Pointersect\cite{pointersect} yields improved performance over NPBG++ \cite{npbg++} and Pointersect.
    }
    \label{tab:comparison_pointersect}
\vspace{-10pt}
\end{table}

\subsection{Ablation Study}
\label{exp:ablation_study}

\begin{figure}[t]
\centering
\includegraphics[width=0.95\linewidth]{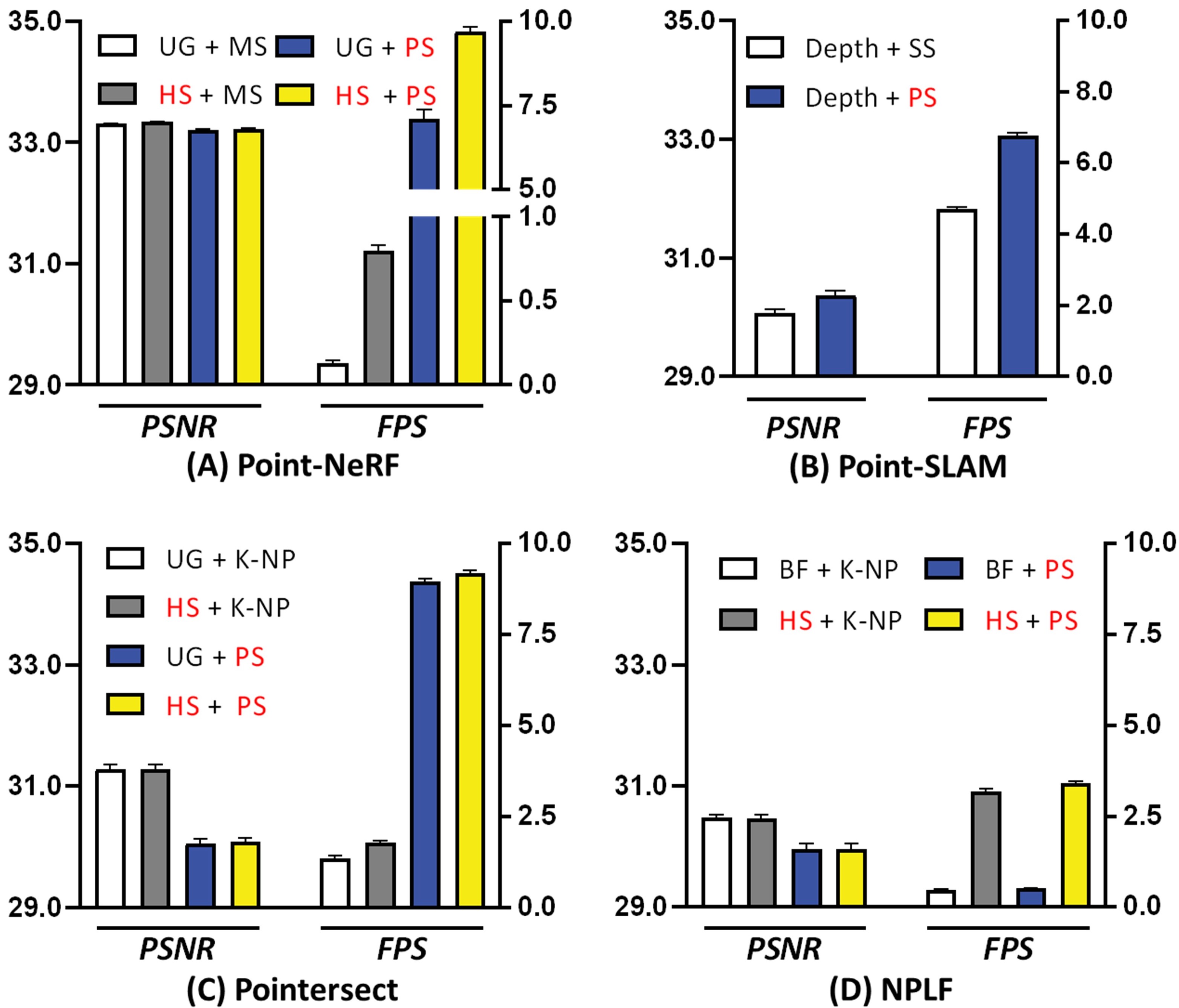} 
\caption{
Ablation study NeRF-Synthesis dataset.
\red{\textit{HS}} and \red{\textit{PS}} represent our \emph{Hashed-Point Searching} and \emph{Adaptive Primary Surface Sampling}.
\textbf{(A)} Combination of \textit{UG} (Uniform Grid) searching with \textit{MS} (Multi-Surfaces) sampling, alongside our techniques.
\textbf{(B)} Combination of \textit{SS} (single surface sampling) with \textit{Our \red{PS}} using \textit{Depth} guided sampling. 
\textbf{(C)} Combination with  \textit{UG} and \textit{K-NP} (\textit{K} Nearest Points) selection for sampling.
\textbf{(D)} Combination with \textit{K-NP} and \textit{BF} (Brute Force) searching.
}
\label{fig:ablation}
\vspace{-8pt}
\end{figure}

\noindent\textbf{Comparison with Point-NeRF~\cite{pointnerf}.} Figure~\ref{fig:ablation}A illustrates the comparison between Point-NeRF's (\textit{UG}) and (\textit{MS}), and our (\textit{HS}) and (\textit{PS}). \textit{HS} notably outperforms \textit{UG} by 5 times in speed with \textit{MS}, while \textit{PS} leads to a dramatic speed increase of 60 to 80 times. \textit{HS} and \textit{PS} both increase efficiency without sacrificing accuracy, with \textit{PS} dominating due to reduced point cloud input for feature extraction and MLP prediction. 
\label{ablation:pointnerf}

\noindent\textbf{Comparison with Point-SLAM~\cite{pointslam}.} Figure~\ref{fig:ablation}B presents the advantage of primary surface sampling (\textit{PS}), which adaptively collects 1 to \textit{n} points on the surface guided by point cloud distribution, over single surface sampling (\textit{SS}), which consistently gather $\hat{n}$ points. The experiment demonstrates that our primary surface sampling not only speeds up the process but also enhances precision. 
\label{ablation:pointslam}

\noindent\textbf{Comparison with Pointersect~\cite{pointersect}.} Figure~\ref{fig:ablation}C,  illustrates that selecting the \textit{K} nearest points (\textit{K-NP}, \textit{K}=40) from noise-initialized point clouds near a camera ray results in higher accuracy compared to using only \textit{PS}. However, our \textit{HS} method proves faster than the \textit{UG}, contributing significantly to the overall process acceleration.
\label{ablation:pointersect}

\noindent\textbf{Comparison with NPLF~\cite{nplf}.} Figure~\ref{fig:ablation}D shows that in noisy point cloud, the \textit{K} nearest points method (\textit{K-NP}) outperforms primary surface sampling (\textit{PS}). Conversely, Figure~\ref{fig:vis_comparison_with_nplf} illustrates that \textit{PS} is superior to \textit{K-NP} in correct point clouds. Additionally, Hashpoint searching proves to be five times faster than the brute force (\textit{BF}) approach. 
\label{ablation:nplf}

\begin{figure}[t]
\centering
\includegraphics[width=0.8\linewidth]{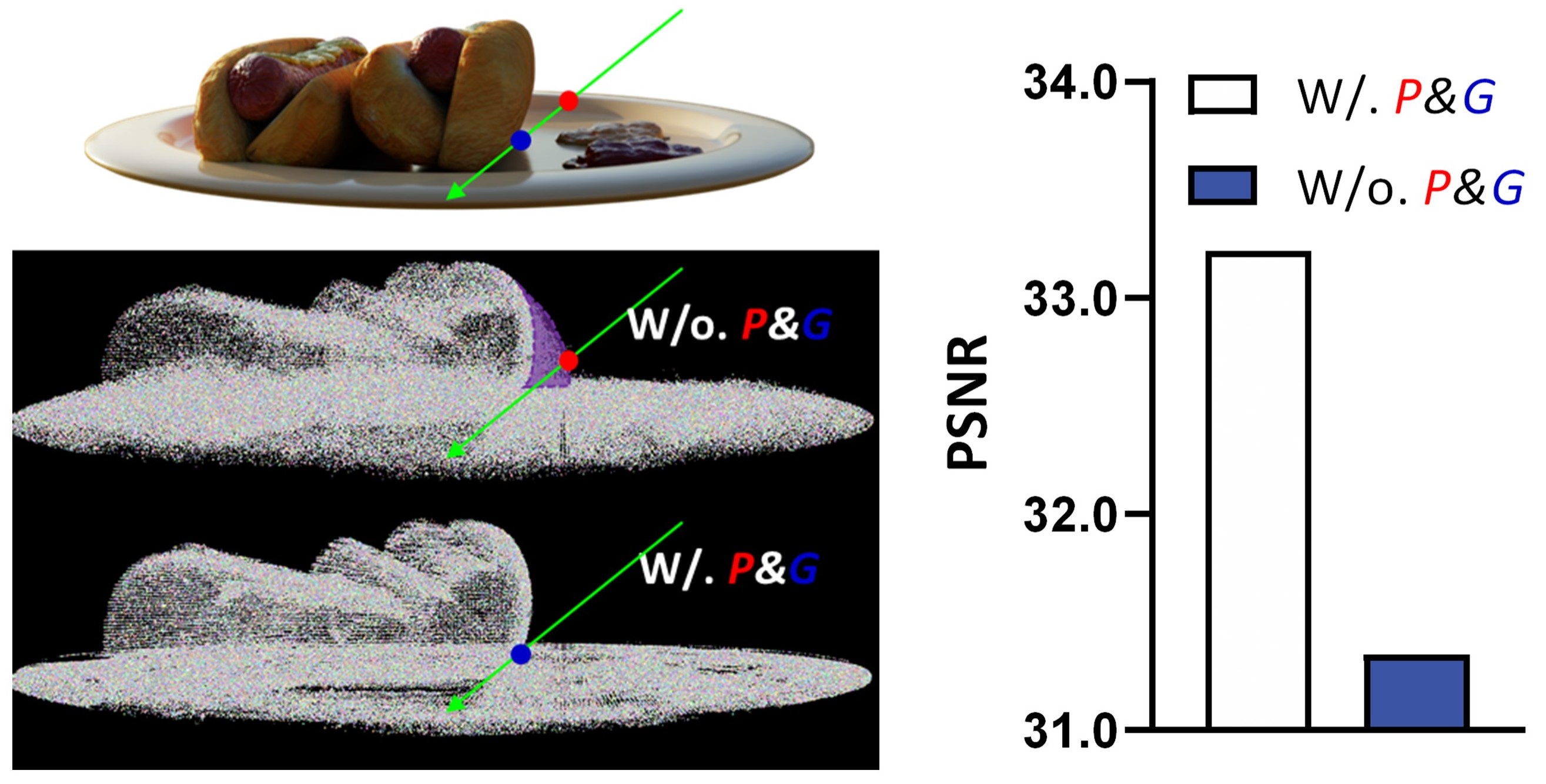} 
\caption{ Illustration of geometry refinement necessity.
\textbf{Left}: Comparison of methods with and without point pruning and growing (\textit{P\&G}) for primary surface sampling. Without \textit{P\&G}, points (in red) are sampled from a noisy surface (in purple), deviating from the true surface (in blue). \textbf{Right}: Performance comparison.}
\label{fig:geo_refine_radius}
\end{figure}

\noindent\textbf{Geometry refinement.} 
\label{ablation:geometry_refine}
In the ablation study, we initialized point clouds with Mvsnet\cite{mvsnet} and followed Point-NeRF's strategy for optimizing noisy points through point pruning and growing (\textit{P\&G}). Geometric optimization was not applied in Point-SLAM, Pointersect, and NPLF, resulting in lower performance. To isolate the algorithm design factors, we test our method with and without refinement on the base of \cite{pointnerf}. Figure~\ref{fig:geo_refine_radius} shows the optimization improves performance, as it mitigates the impact of geometric noise on feature extraction from primary surfaces. Despite relying on geometric structures, it is easily obtained with a brief 10-minute point optimization during training, enabling significant speed gains without losing accuracy. We also explored the combination of $\gamma$ and $\beta$ parameters to manage noisy input. Further details are available in our Supplementary.

\section{Conclusion}
Our method enhances point cloud rendering speed by combining ray tracing with rasterization. Using our HashPoint technique, point clouds are efficiently organized in a hash table through rasterization, accelerating searches. This approach, coupled with primary surface sampling, reduces input points and leverages geometric distribution, significantly speeding up rendering. For instance, searching through a million points takes only 0.5 ms on a standard GPU. In selection, our approach outperforms \textit{K} nearest point methods in accuracy and is faster than multi-surface sampling. Easily integrated with existing methods, HashPoint advances accuracy and rendering speed, pushing the boundaries of point cloud rendering.

{
     \small
     \bibliographystyle{ieeenat_fullname}
     \bibliography{main}
}

\clearpage
\setcounter{page}{1}
\maketitlesupplementary


\setcounter{section}{0}

\noindent In this supplementary material, we provide details about the following topics:
\begin{itemize}
    \item \textit{Derivation of searching radius} in Appendix~\ref{sup:derivation_r};
    \item \textit{Additional experiments} in Appendix~\ref{sup:addtional_exp};
    \item \textit{Breakdown results} in Appendix~\ref{sup:breakdown_results};
    \item \textit{Visualization} in Appendix~\ref{sup:visualization};
    \item \textit{Limitation} in Appendix~\ref{sup:limitation}.
\end{itemize}

\begin{figure}[t!]
\centering
\includegraphics[width=1.0\linewidth]{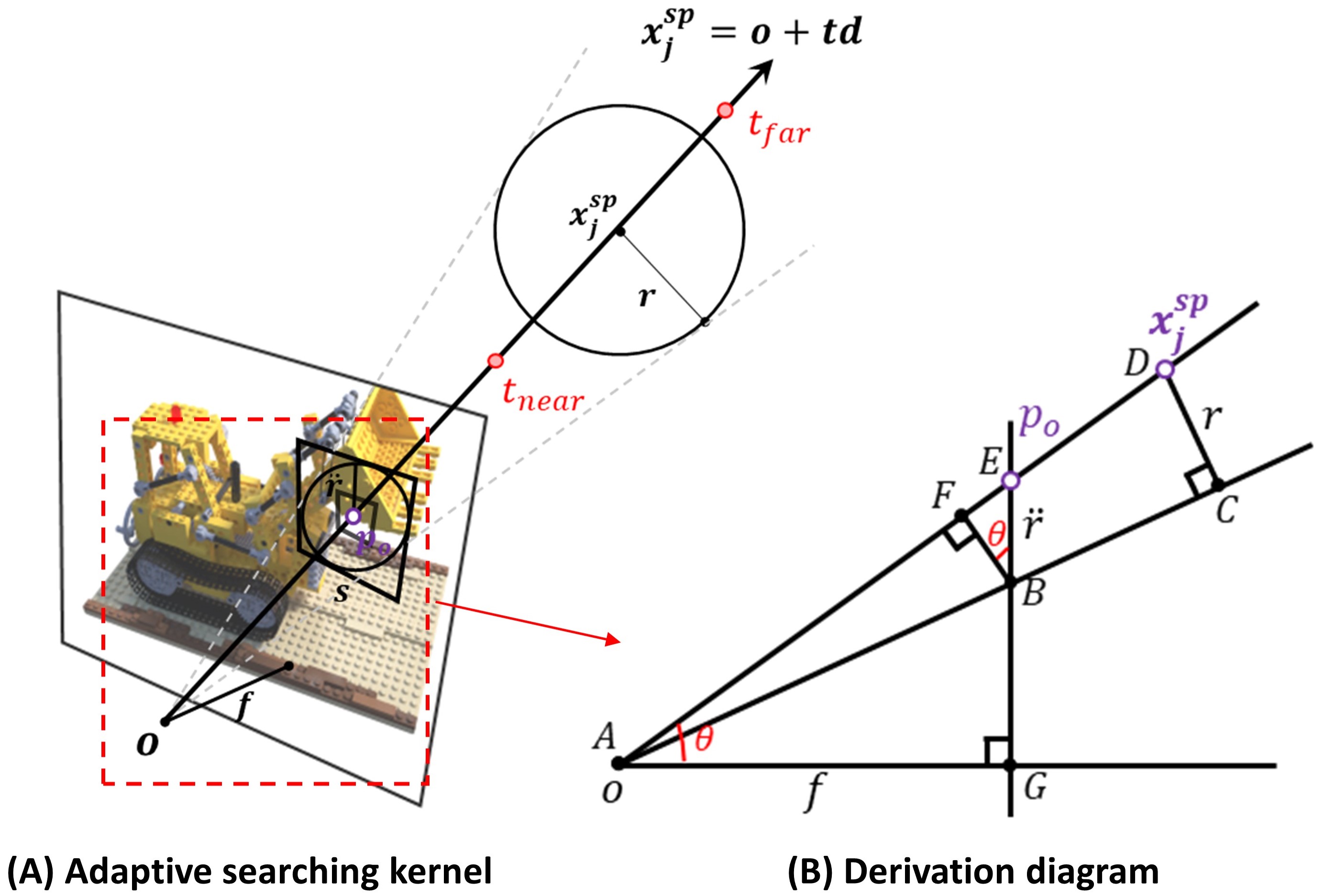}
\caption{ (A) Adaptive searching kernel applied on an image plane. The square represents a pixel to be rendered on the image. The circle represents a cross-section of the searching kernel of a cone to select points to render. (B) Detailed derivation of the searching kernel.}
\vspace{-8pt}
\label{fig:adaptive_sampling_derivation}
\end{figure}

\vspace{-6pt}
\section{Adaptive searching radius derivations}
\label{sup:derivation_r}
In this section, we elaborate on the derivation of adaptive searching kernel $r$ as mentioned in Section~\ref{Method:point_searching}. We initiate the derivation with a disc representation on the image plane, approximating the area of the pixel. The radius of the disc can be calculated by  $\dot{r}=\sqrt{\Delta x \cdot \Delta y / \pi}$, where $\Delta x$ and $\Delta_y$ are the width and height of the pixel in world coordinates. For a broader coverage area, we shift from a disc with radius $\dot{r}$ for one pixel to a larger disc with radius $\ddot{r}$ for the searching kernel. The searching kernel size $s$ can be expressed as $s = 2 \cdot \left\lceil \frac{\ddot{r}}{\dot{r}} \right\rceil + 1$. Specifically, given the searching kernel $s$ and radius $\dot{r}$, the suitable range for $\ddot{r}$ is,
%
\begin{equation}
\begin{aligned}
\left(\frac{s - 1}{2} - 1\right) \cdot \dot{r} < \ddot{r} \leq \left(\frac{s - 1}{2}\right) \cdot \dot{r} .
\end{aligned}
\end{equation}

\noindent As shown in Figure~\ref{fig:adaptive_sampling_derivation}B, to determine the radius $r$ for sample points, we employ the concept of similar triangles, specifically $\triangle ABF \sim \triangle ADC$. This set of similar triangles establishes a proportionality that allows us to solve for $r$ as,
%
\begin{equation}
\begin{aligned}
r = CD = \frac{AD \times BF}{AB} .
\end{aligned}
\label{eq:r_original}
\end{equation}

\begin{figure}[t!]
\centering
\includegraphics[width=1.0\linewidth]{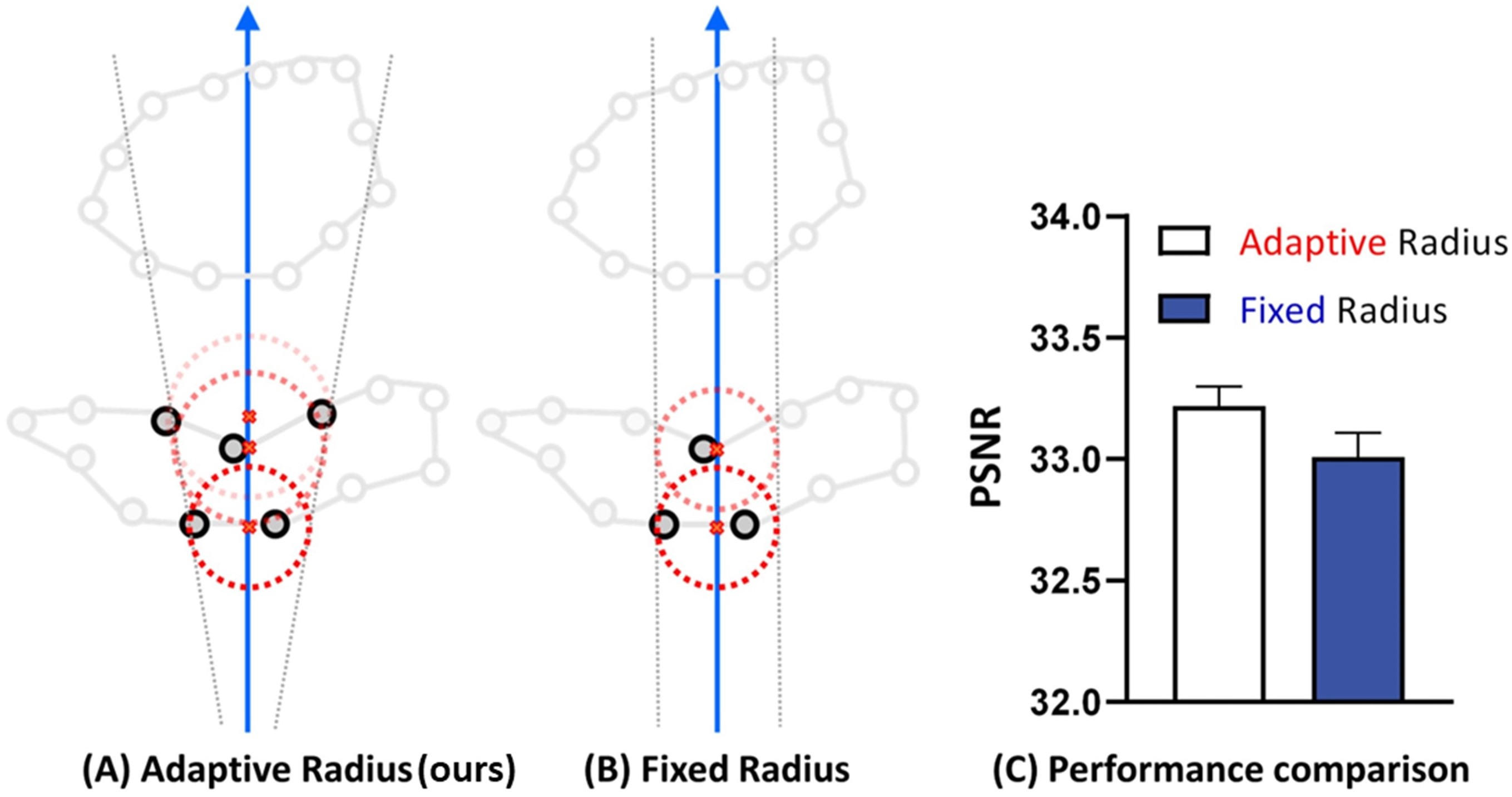}
\caption{ Radius Strategies and Performance: (A) depicts an adaptive radius correlated with viewport distance, (B) shows a constant fixed radius, and (C) compares their PSNR performance. Additional comparison of point cloud selection strategies in Figure~\ref{fig:sampling} in the main text.}
\vspace{-6pt}
\label{fig:radius_comparison}
\end{figure}

\noindent Here, $AD$ represents the Euclidean distance between the camera ray $\mathbf{o}$ and the sample point position $\mathbf{x_{j}^{sp}}$,given by $AD = \left\| \mathbf{x_{j}^{sp}} - \mathbf{o}\right\|_2$. By invoking another set of similar triangles, $\triangle BEF \sim \triangle AEG$, we can calculate $BF$ as,
%
\begin{equation}
\begin{aligned}
BF = \ddot{r} \cdot cos{\theta} 
   = \frac{\ddot{r} \cdot AG}{AE}
   = \frac{\ddot{r}f}{ \left\| \mathbf{p_o} -\mathbf{o}\right\|_2},
\end{aligned}
\end{equation}

\noindent where $f$ is the local length and $\mathbf{p_o}$ is the pixel center intersected by the camera ray. The length $AB$ is the hypotenuse of the right triangle $\triangle AGB$ and is computed using the Pythagorean theorem as,
%
\begin{equation}
\begin{aligned}
AB &= \sqrt{GB^2 + AG^2} \\
   &= \sqrt{\left(GE - BE\right)^2 + AG^2} \\
   &= \sqrt{\left(\sqrt{AE^2 - AG^2} - BE\right)^2 + AG^2} \\
   &= \sqrt{ \left(\sqrt{ \left\| \mathbf{p_o} -\mathbf{o}\right\|_2^2 - f^2} - \ddot{r}\right)^2 + f^2 }
\end{aligned}
\end{equation}

\begin{figure*}[t!]
\centering
\includegraphics[width=0.96\linewidth]{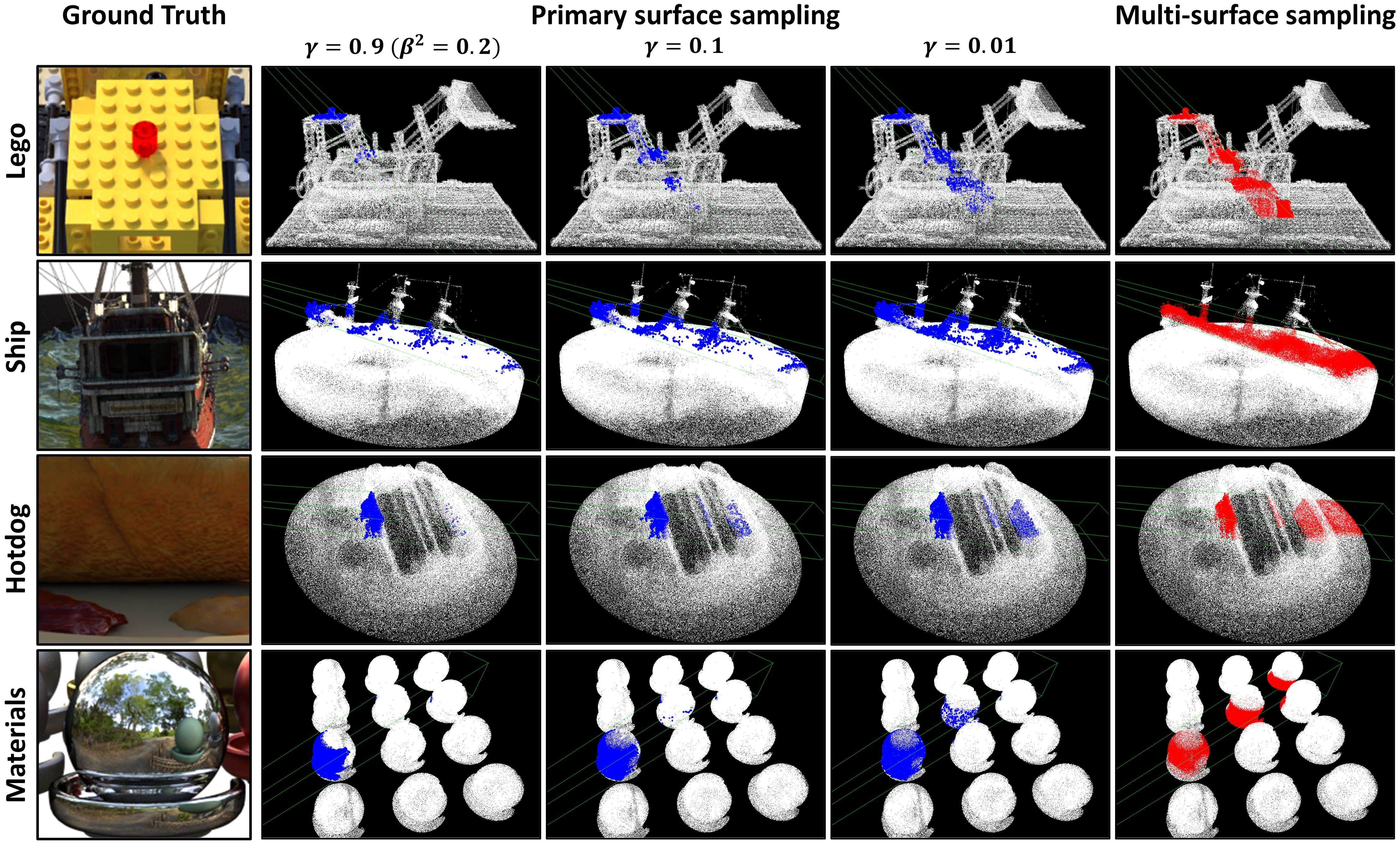}
\caption{Adaptive sampling for noisy input, showing the transition from primary-surface to multiple-surface sampling by adjusting $\gamma$ values, with visual results for different surfaces. It features a visual comparison across different surfaces. The first column displays the ground truth, and the last column showcases the region where multi-surface sampling is applied in Point-NeRF. The middle part illustrates the gradual transition. In the main text, the default value of $\beta^2$ is set to 0.02 when dealing with refined geometry input.}
\label{fig:noisy_input}
\vspace{-6pt}
\end{figure*}

\noindent Combining these elements, the final expression for the radius $r$ of the searching kernel is,
%

\begin{equation}
\begin{aligned}
    r = \frac{\|\mathbf{x^{sp}_j}-\mathbf{o}\|_2 \cdot f \ddot{r}}{\|\mathbf{p_o} - \mathbf{o}\|_2 \cdot \sqrt{\left(\sqrt{\|\mathbf{p_o} - \mathbf{o}\|_2^2-f^2}- \ddot{r}\right)^2+f^2}.} 
\label{eq:radius}
\end{aligned}
\end{equation}

\noindent This integrated expression encapsulates the geometric relationship between the image plane, the searching kernel, and the camera parameters, completing the derivation of the adaptive searching kernel's radius.

\textbf{Room for improvement}. As the angle $\widehat{BOE}$ is minuscule (less than 1 degree), 
we can make the below approximation and then apply a similar triangle relationship:
\vspace{-6pt}
\begin{equation}
\begin{aligned}
AB &\approx AF = AE - EF = AE - \frac{BF \times EG }{AG} \\
\end{aligned}
\label{eq:AB_approx}
\end{equation}



In fact, we can further simplify eq. \ref{eq:AB_approx} by applying small angle assumption of $\widehat{BOE}$ twice: $AB \approx AF \approx AE$. Eq. \ref{eq:r_original} becomes:
\vspace{-6pt}
\begin{equation}
\begin{aligned}
r   &\approx \frac{AD \times BF}{AE} = \frac{\|\mathbf{x^{sp}_j}-\mathbf{o}\|_2 \cdot \frac{\ddot{r} f}{\|\mathbf{p_o} - \mathbf{o}\|_2}}
    {\|\mathbf{p_o} - \mathbf{o}\|_2} \\
   &\approx \frac{\|\mathbf{x^{sp}_j}-\mathbf{o}\|_2 \cdot \ddot{r} f}
    {\|\mathbf{p_o} - \mathbf{o}\|_2^2}
\end{aligned}
\label{eq:r_approx2}
\end{equation}

Eq. \ref{eq:r_approx2} is simpler and faster to computer than eq. \ref{eq:radius}.

\section{Additional experiments}
\label{sup:addtional_exp}

\noindent\textbf{Comparison between the fixed radius and our adaptive radius sampling.} 
Regarding the discussion in Section~\ref{sec:Method}, we 
compared the performance differences between adaptive radius and fixed radius on the NeRF-Synthesis dataset, building upon the Point-NeRF framework. Figure~\ref{fig:radius_comparison} illustrates the differences between adaptive and fixed radius strategies for point cloud sampling in neural radiance fields. As explained in Section~\ref{sec:discussion}, the adaptive radius expands with increased distance from the ray origin, resembling a camera's view frustum, while the fixed radius used by Point-NeRF remains constant. The performance in Figure~\ref{fig:radius_comparison}C demonstrates that the adaptive radius yields higher PSNR values, indicating a reduction in noise by selecting only points relevant to the rendered pixel.

\begin{table*}[!t]
    \centering
    \resizebox{0.76\textwidth}{!}{
    \begin{tabular}{lccccccccc}
    \toprule
    \textbf{Method} & \texttt{Rm\,0} & \texttt{Rm\,1} & \texttt{Rm\,2} & \texttt{Off\,0} & \texttt{Off\,1} & \texttt{Off\,2} & \texttt{Off\,3} & \texttt{Off\,4} & Avg. \\ \hline
    NICE-SLAM \cite{niceslam} & 0.97 & 1.31 & 1.07 & 0.88 & 1.00 & 1.06 & 1.10 & 1.13 & 1.06 \\
    Point-SLAM \cite{pointslam} & \textbf{0.61} & \textbf{0.41} & \textbf{0.37} & \textbf{0.38} & {0.48} & \textbf{0.54} & {0.69} & {0.72} & \textbf{0.52} \\ 
    Point-SLAM + Ours & 0.69 & 0.53 & \textbf{0.37} & 0.47 & \textbf{0.45} & 0.65 & \textbf{0.64} & \textbf{0.54} & 0.54\\ 
    \bottomrule
    \end{tabular}}
    \vspace{-3pt}
    \caption{Tracking performance on Replica~\cite{replica} (ATE
RMSE $\downarrow$ [cm]).}
    \label{tab:tracking_replica}
\end{table*}

\begin{table*}[!t]
\centering
\resizebox{0.92\textwidth}{!}{
\begin{tabular}{llccccccccc}
\hline
\textbf{Method} & \textbf{Metric} & \texttt{Rm\,0} & \texttt{Rm\,1} & \texttt{Rm\,2} & \texttt{Off\,0} & \texttt{Off\,1} & \texttt{Off\,2} & \texttt{Off\,3} & \texttt{Off\,4} & Avg. \\ \hline
\multirow{4}{*}{NICE-SLAM\cite{niceslam}} & Depth L1 [cm] $\downarrow$ & 1.81 & 1.44 & 2.04 & 1.39 & 1.76 & 8.33 & 2.01 & 2.97 & 2.97 \\
                                          & Precision [\%] $\uparrow$ & 45.86 & 43.76 & 44.38 & 51.40 & 50.80 & 38.37 & 40.85 & 37.35 & 44.10 \\
                                          & Recall [\%] $\uparrow$ & 44.10 & 46.12 & 42.78 & 48.66 & 53.08 & 39.98 & 39.24 & 35.77 & 43.69 \\
                                          & F1 [\%] $\uparrow$ & 44.96 & 44.84 & 43.56 & 49.99 & 51.91 & 39.16 & 39.92 & 36.54 & 43.86 \\
\hline
\multirow{4}{*}{Point-SLAM\cite{pointslam}} & Depth L1 [cm] $\downarrow$ & 0.53 & \textbf{0.22} & \textbf{0.46} & \textbf{0.30} & 0.57 & \textbf{0.49} & \textbf{0.51} & 0.46 & \textbf{0.44}\\
                                            & Precision [\%] $\uparrow$ & 91.95 &\textbf{99.04} &\textbf{97.89} & \textbf{99.00} & \textbf{99.37} &98.05 &96.61 &\textbf{93.98} &96.99 \\
                                            & Recall [\%] $\uparrow$ & 82.48 &\textbf{86.43} &84.64 &\textbf{89.06} &84.99 &81.44 &81.17 &78.51 &83.59 \\ 
                                            & F1 [\%] $\uparrow$ & 86.90 &\textbf{92.31}& \textbf{90.78} & \textbf{93.77} &91.62 &88.98& 88.22& 85.55 &89.77 \\
\hline
\multirow{4}{*}{Point-SLAM + Ours} & Depth L1 [cm] $\downarrow$ & \textbf{0.51} & 0.23 & 0.55 & 0.37 & \textbf{0.46} & 0.54 & 0.52 & \textbf{0.42} & 0.45\\
                                            & Precision [\%] $\uparrow$  & \textbf{92.31} & 98.96 & 97.56 & 98.85 & 99.31 & \textbf{99.33} & \textbf{97.45} & 93.95 &\textbf{97.22} \\
                                            & Recall [\%] $\uparrow$  & \textbf{83.32} & 85.98 & \textbf{84.65} & 88.86 & \textbf{85.23} & \textbf{81.74} & \textbf{82.71} & \textbf{79.38} & \textbf{83.98}\\ 
                                            & F1 [\%] $\uparrow$ & \textbf{87.58} & 92.01 & 90.64 & 93.58 & \textbf{91.73} & \textbf{89.68} & \textbf{89.38} & \textbf{86.05} & \textbf{90.11}\\
\bottomrule                                
\end{tabular}}
\vspace{-3pt}
\caption{Reconstruction performance on Replica~\cite{replica}.}
\vspace{-6pt}
\label{tab:reconstruction_replica}
\end{table*}

\begin{figure}[t!]
\centering
\includegraphics[width=0.98\linewidth]{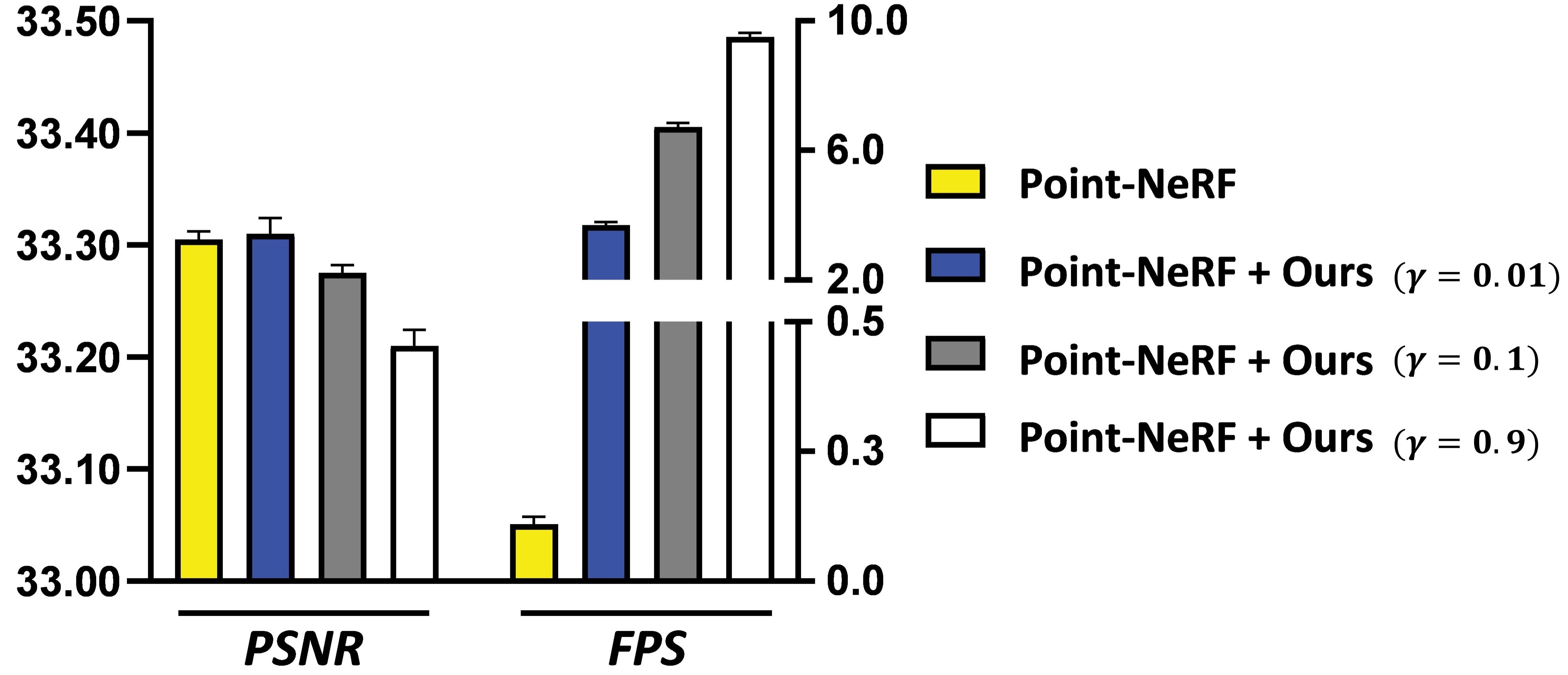}
\caption{ The illustration of the trade-off between efficiency and accuracy in the transition from primary surface to multiple surface sampling under the noisy input by adjusting the $\gamma$ parameter. Point-NeRF with our HashPoint and $\gamma=0.01$ outperforms original Point-NeRF in both \emph{PSNR} and \emph{FPS}.}
\vspace{-9pt}
\label{fig:noisy_input_exp}
\end{figure}

\noindent\textbf{Our adaptive sampling for noisy input.} 
This section extends the discussion on the solution to noisy input from Section~\ref{ablation:geometry_refine}. Primary surface sampling encounters challenges in accurately extracting surface features from noisy point cloud input. As outlined in the ablation study, we proposed two methods to address issues with noisy inputs. The first involves a brief 10-minute geometry optimization that significantly enhances the benefit of our methods by filling gaps and refining noisy point clouds from noisy surfaces. The second method adaptively adjusts the sampling scope, shifting from primary surface to multi-surface sampling. Notably, unlike full multi-surface sampling, which gathers features from all intersected surfaces, our approach, as illustrated in Figure~\ref{fig:noisy_input}, samples just slightly beyond the primary surface, avoiding extensive multi-surface feature collection. 
Figure \ref{fig:noisy_input_exp} demonstrates that reducing $\gamma$ increases precision and reduces the speed of sampling. A larger $\gamma$ (\textit{e.g.}, 0.9) focuses on primary surface sampling, maximizing speed for interactive frame rate while maintaining competitive precision. Conversely, a smaller $\gamma$ captures more point cloud features below the primary surface, exceeding the precision of multi-surface sampling and still achieving speeds over 20 times faster. In summary, the $\gamma$ adjustment effectively balances sampling speed and precision, with each setting offering unique benefits in efficiency and accuracy.
\label{sup:adaptive_sampling}

\begin{table}[!t]
    \centering
    \resizebox{0.4\textwidth}{!}{
    \begin{tabular}{cccc}
    \toprule
    \multicolumn{2}{c}{\textbf{Method}} & \multicolumn{2}{c}{\textbf{Performance}} \\
    \midrule
        Searching & Selection & PSNR $\uparrow$  & FPS $\uparrow$ \\
    \midrule
    \multicolumn{4}{c}{Point-NeRF}  \\
    \midrule
        Uniform grid & Multiple surface & 33.31 & 0.12\\
        HashPoint & Multiple surface & 33.34 & 0.78\\
        Uniform grid & Primary surface & 33.20 & 6.92\\
        HashPoint & Primary surface & 33.22 & 9.60\\
    \midrule
    \multicolumn{4}{c}{Point-SLAM}  \\
    \midrule
        Depth & Single surface & 30.01& 4.64\\
        Depth & Primary surface & 30.32& 6.72\\
   
    \midrule
    \multicolumn{4}{c}{Pointersect}  \\
    \midrule
        Uniform grid & \textit{K} nearest points & 31.23& 1.30\\  
        HashPoint & \textit{K} nearest points & 31.23 & 1.75\\
        Uniform grid & Primary surface & 30.01& 8.89\\ 
        HashPoint & Primary surface & 30.03& 9.13\\
    \midrule
    \multicolumn{4}{c}{NPLF}  \\
    \midrule
        Brute force & \textit{K} nearest points & 30.44 & 0.48\\
        HashPoint & \textit{K} nearest points & 30.42& 3.12\\
        Brute force & Primary surface & 29.88 & 0.50\\
        HashPoint & Primary surface & 29.90& 3.40\\
    
    \bottomrule
    \end{tabular}}
    \vspace{-3pt}
    \caption{Ablation study evaluates on NeRF-Synthesis dataset. }
    \label{tab:ablation}
    \vspace{-10pt}
\end{table}

\begin{table*}[!t]
    \centering
    \resizebox{0.78\textwidth}{!}{
    \begin{tabular}{lcccccccc}
        \toprule
        & Chair & Drums & Lego & Mic & Materials & Ship & Hotdog & Ficus \\
        \midrule
        \multicolumn{9}{c}{PSNR$\uparrow$} \\
        \midrule
        NPBG  & 26.47 & 21.53 & 24.84 & 26.62 & 21.58 & 21.83 & 29.01 & 24.60 \\
        NeRF  & 33.00 & 25.01 & 32.54 & 32.91 & 29.62 & 28.65 & 36.18 & 30.13 \\
        NSVF  & 33.19 & 25.18 & 32.54 & 34.21 & 29.62 & 27.93 & 37.14 & 31.23 \\
        Point-NeRF & 35.40 & 26.06 & \textbf{35.04} & 35.95 & 29.61 & 30.97 & \textbf{37.30} & \textbf{36.13} \\
        Point-NeRF + Ours & \textbf{35.54} & \textbf{26.12} & 34.68 & \textbf{36.34} & \textbf{30.64} & \textbf{31.08} & 37.02 & 34.30 \\ 
        \midrule
        \multicolumn{9}{c}{SSIM$\uparrow$} \\
        \midrule
        NPBG & 0.939 & 0.904 & 0.923 & 0.959 & 0.887 & 0.866 & 0.964 & 0.940 \\
        NeRF & 0.967 & 0.925 & 0.961 & 0.980 & 0.949 & 0.856 & 0.974 & 0.964 \\
        NSVF & 0.968 & 0.931 & 0.960 & \textbf{0.987} & \textbf{0.973} & 0.854 & 0.980 & 0.973 \\
        Point-NeRF & \textbf{0.984} & \textbf{0.935} & \textbf{0.978} & \textbf{0.990} & 0.948 & 0.892 & \textbf{0.982} & 0.987 \\
        Point-NeRF + Ours & 0.977 & 0.931 & 0.967 & 0.984 & 0.949 & \textbf{0.920} & 0.978 & 0.978 \\ 
        \midrule
        \multicolumn{9}{c}{LPIPS$_{vgg}\downarrow$} \\
        \midrule
        NPBG & 0.085 & 0.112 & 0.119 & 0.060 & 0.134 & 0.210 & 0.075 & \textbf{0.078} \\
        NeRF & 0.046 & 0.091 & 0.050 & 0.028 & \textbf{0.063} & 0.206 & 0.121 & 0.044 \\
        Point-NeRF & \textbf{0.023} & \textbf{0.078} & \textbf{0.024} & \textbf{0.014} & 0.072 & 0.124 & 0.037 & \textbf{0.022} \\
        Point-NeRF + Ours & 0.028 & 0.101 & 0.047 & 0.018 & 0.075 & \textbf{0.097} & \textbf{0.036} & 0.041 \\ 
        \midrule
        \multicolumn{9}{c}{LPIPS$_{alex}\downarrow$} \\
        \midrule
        NSVF & 0.043 & 0.069 & 0.029 & 0.010 & \textbf{0.021} & 0.162 & 0.025 & 0.017 \\
        Point-NeRF & \textbf{0.010} & \textbf{0.055} & \textbf{0.011} & \textbf{0.007} & 0.041 & 0.070 & \textbf{0.016} & \textbf{0.009} \\
        Point-NeRF + Ours & 0.012 & 0.067 & 0.014 & 0.010 & 0.049 & \textbf{0.054} & 0.018 & 0.017 \\ 
        \bottomrule
    \end{tabular}}
    \caption{Quantitative results in the NeRF Synthetic dataset.}
    \vspace{-4pt}
    \label{tab:nerf_synthetic}
\end{table*}

\begin{figure*}[t!]
\centering
\includegraphics[width=0.99\linewidth]{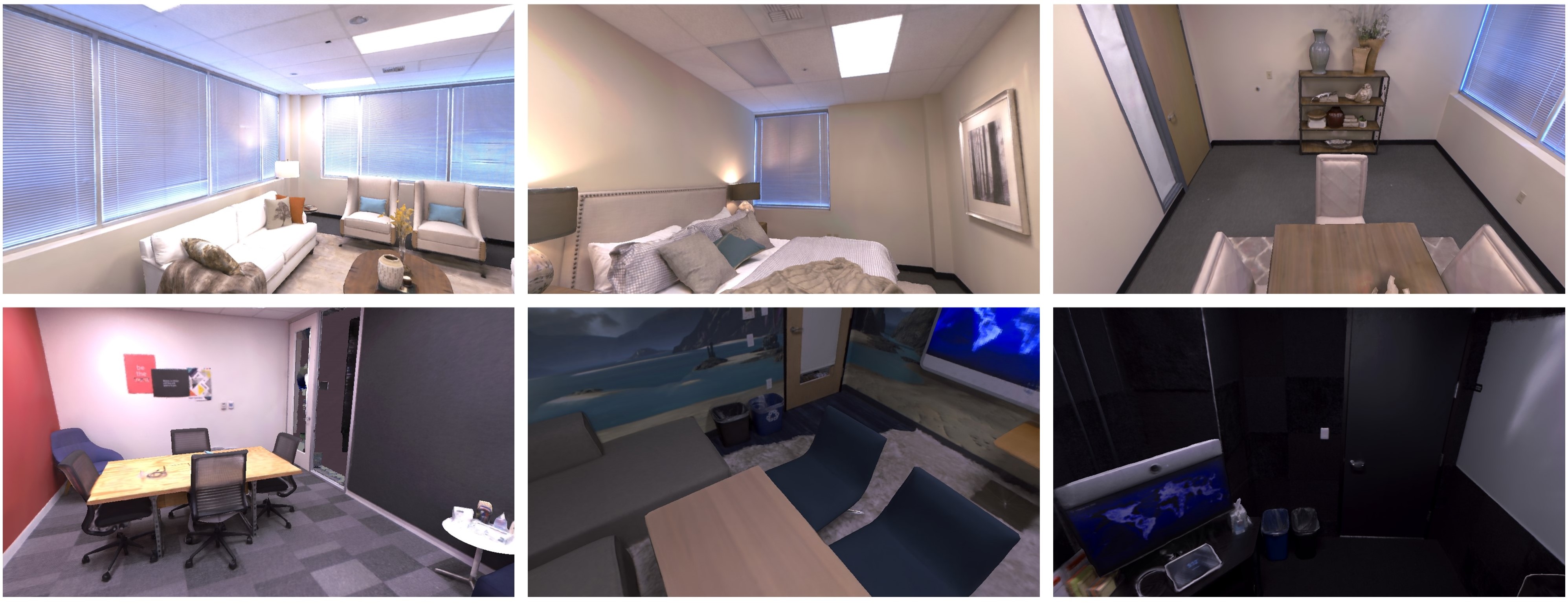}
\caption{ The qualitative results of Point-SLAM~\cite{pointnerf} with ours on the Replica~\cite{replica} dataset. }
\vspace{-9pt}
\label{fig:replica}
\end{figure*}

\noindent\textbf{Tracking and mapping.} 
We supplement the integration of our method with Point-SLAM~\cite{pointslam} for tracking and reconstruction on the Replica~\cite{replica} dataset. As shown in Tables~\ref{tab:tracking_replica} and~\ref{tab:reconstruction_replica}, our approach outperforms the original depth-guided sampling in terms of precision and recall, while maintaining competitive performance in other metrics.

\section{Breakdown results}
\label{sup:breakdown_results}
Table~\ref{tab:nerf_synthetic} provides detailed per-scene quantitative results comparing integration with Point-NeRF on the NeRF-Synthesis dataset. Our integration significantly accelerates the process while preserving competitive performance. The comparative data for our ablation study is outlined in Table~\ref{tab:ablation}.

\section{Visualization}
\label{sup:visualization}
\begin{figure*}[t!]
\centering
\includegraphics[width=0.95\linewidth]{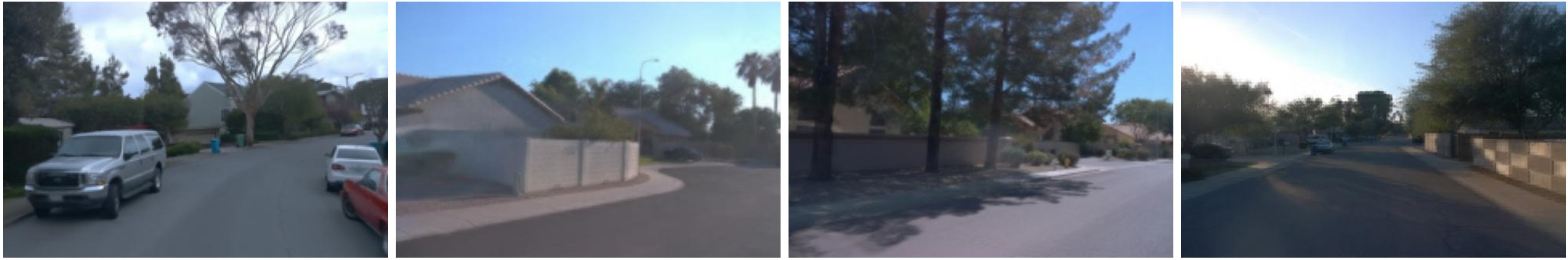}
\caption{ The qualitative results of NPLF~\cite{nplf} with ours on the Waymo~\cite{waymo} dataset. }
\vspace{-9pt}
\label{fig:waymo}
\end{figure*}

We also experiment with the integration with Point-SLAM and NPLF on Replica and Waymo datasets respectively. The qualitative results are shown in Figure~\ref{fig:replica} and~\ref{fig:waymo}. Please find more visual results for sampling comparison in our video.

\section{Limitation}
\label{sup:limitation}
During optimization, due to gradient propagation issues, multi-surface sampling is still necessary to sample and optimize all points as much as possible. Currently, our choice of $\beta$ is fixed and does not dynamically adjust based on the geometry's distribution and noise level. Future work could explore dynamic adjustment of the sampling process. For instance, as the geometry is progressively optimized, $\beta$ could increase gradually, transitioning from multi-surface to primary surface sampling.


\end{document}